# Deep Semi-Random Features for Nonlinear Function Approximation

**Kenji Kawaguchi,[1*] Bo Xie,[2*] Vikas Verma,[3] Le Song[2]**
[1]Massachusetts Institute of Technology
[2]Georgia Institute of Technology
[3]Aalto University

**Abstract**

We propose semi-random features for nonlinear function approximation. The flexibility of semi-random feature lies between the fully adjustable units in deep learning and the random features used in kernel methods. For one hidden layer models with semi-random features, we prove with no unrealistic assumptions that the model classes contain an arbitrarily good function as the width increases (universality), and despite non-convexity, we can find such a good function (optimization theory) that generalizes to unseen new data (generalization bound). For deep models, with no unrealistic assumptions, we prove universal approximation ability, a lower bound on approximation error, a partial optimization guarantee, and a generalization bound. Depending on the problems, the generalization bound of deep semi-random features can be exponentially better than the known bounds of deep ReLU nets; our generalization error bound can be independent of the depth, the number of trainable weights as well as the input dimensionality. In experiments, we show that semi-random features can match the performance of neural networks by using slightly more units, and it outperforms random features by using significantly fewer units. Moreover, we introduce a new implicit ensemble method by using semi-random features.

## Introduction

Many recent advances, such as human-level image classification (Deng et al. 2009) and game playing (Bellemare et al. 2013; Silver et al. 2016) in machine learning are attributed to large-scale nonlinear function models. There are two dominating paradigms for nonlinear modeling in machine learning: kernel methods and neural networks:

- Kernel methods employ *pre-defined* basis functions, $k(x, x')$, called kernels to represent nonlinear functions (Scholkopf and Smola 2001; Shawe-Taylor and Cristianini 2004). Learning algorithms that use kernel methods often come with nice theoretical properties–globally optimal parameters can be found via convex optimization, and statistical guarantees can be provided rigorously. However, kernel methods typically work with matrices that are quadratic in the number of samples, leading to unfavorable computation and storage complexities. A popular approach to tackle such issues is to approximate kernel functions using random features (Rahimi and Recht 2008; Sindhwani, Avron, and Mahoney 2014; Pennington, Yu, and Kumar 2015). One drawback of random features is that its approximation powers suffer from the curse of dimensionality (Barron 1993) because its bases are not adaptive to the data.
- Neural networks use adjustable basis functions and learn their parameters to approximate the target nonlinear function (LeCun, Bengio, and Hinton 2015). Such adaptive nature allows neural networks to be compact yet expressive. As a result, they can be efficiently trained on some of the largest datasets today. By incorporating domain specific network architectures, neural networks have also achieved state-of-the-art results in many applications. However, learning the basis functions involves difficult non-convex optimization. Few theoretical insights are available in the literature and more research is needed to understand the working mechanisms and theoretical guarantees for neural networks (Choromanska, LeCun, and Arous 2015; Swirszcz, Czarnecki, and Pascanu 2016; Shamir 2016).

Can we have the best of both worlds? Can we develop a framework for big nonlinear problems which has the ability to adapt basis functions, has low computational and storage complexity, while at the same time retaining some of the theoretical properties of random features? Towards this goal, we propose semi-random features to explore the space of trade-off between flexibility, provability and efficiency in nonlinear function approximation. We show that semi-random features have a set of nice theoretical properties, like random features, while possessing a (deep) representation learning ability, like deep learning. More specifically:

- Despite the nonconvex learning problem, semi-random feature model with one hidden layer has no bad local minimum;
- Depending on the problems, the generalization bound of deep semi-random features can be exponentially better than known bounds of deep ReLU nets;
- Semi-random features can be composed into multi-layer architectures, and going deep in the architecture leads to more expressive model than going wide;
- Semi-random features also lead to statistical stable func-

[*]Equal contribution

tion classes, where generalization bounds can be readily provided.

## Background

We briefly review different ways of representing nonlinear functions in this section.

**Hand-designed basis.** In a classical machine learning approach for nonlinear function approximation, users or domain experts typically handcraft a set of features $\phi_{\text{expert}} : \mathcal{X} \to \mathcal{H}$, a map from an input data space $\mathcal{X}$ to a (complete) inner product space $\mathcal{H}$. Many empirical risk minimization algorithms then require us to compute the inner product of the features as $\langle \phi_{\text{expert}}(x), \phi_{\text{expert}}(x') \rangle_{\mathcal{H}}$ for each pair $(x, x') \in \mathcal{X} \times \mathcal{X}$. Computing this inner product can be expensive when the dimensionality of $\mathcal{H}$ is large, and indeed it can be infinite. For example, if $\mathcal{H}$ is the space of square integrable functions, we need to evaluate the integral as $\langle \phi_{\text{expert}}(x), \phi_{\text{expert}}(x') \rangle_{\mathcal{H}} = \int_{\omega} \phi_{\text{expert}}(x;\omega)\overline{\phi_{\text{expert}}(x';\omega)}$.

**Kernel methods.** When our algorithms solely depend on the inner product, the kernel trick avoids this computational burden by introducing an easily computable kernel function as $k_{\text{expert}}(x', x) = \langle \phi_{\text{expert}}(x), \phi_{\text{expert}}(x') \rangle_{\mathcal{H}}$, resulting in an implicit definition of the features $\phi_{\text{expert}}$ (Scholkopf and Smola 2001; Shawe-Taylor and Cristianini 2004). However, the kernel approach typically scales poorly on large datasets. Given a training set of $m$ input points $\{x_i\}_{i=1}^m$, evaluating a learned function at a new point $x$ requires computing $\hat{f}(x) = \sum_{i=1}^m \alpha_i k_{\text{expert}}(x_i, x)$, the cost of which increases linearly with $m$. Moreover, it usually requires computing (or approximating) inverses of matrices of size $m \times m$.

**Random features.** In order to scale to large datasets, one can approximate the kernel by a set of random basis functions sampled according to some distributions. That is, $k_{\text{expert}}(x', x) \approx \frac{1}{C}\sum_{j=1}^C \phi_{\text{random}}(x; \mathbf{r}_j)\phi_{\text{random}}(x'; \mathbf{r}_j)$, where both the type of basis functions $\phi_{\text{random}}$, and the sampling distribution for the random parameter $\mathbf{r}_j$ are determined by the kernel function. Due to its computational advantage and theoretical foundation, the random feature approach has many applications and is an active research topic (Rahimi and Recht 2008; Sindhwani, Avron, and Mahoney 2014; Pennington, Yu, and Kumar 2015).

**Neural networks.** Neural networks approximate functions using weighted combination of *adaptable* basis functions $f(x) = \sum_{k=1}^n w_k^{(2)} \phi(x; \mathbf{w}_k^{(1)})$, where both the combination weights $w_k^{(2)}$ and the parameters $\mathbf{w}_k^{(1)}$ in each basis function $\phi$ are learned from data. Neural networks can be composed into multilayers to express highly flexible nonlinear functions.

## Semi-Random Features

When comparing different nonlinear representaitons, we can see that random features are designed to approximate a known kernel, but not for learning these features from the given dataset (i.e., it is not a *representation learning*). As a result, when compared to neural network, it utilizes less amount of information encoded in the dataset, which could be disadvantageous. Neural networks, on the other hand, pose a difficulty for theoretical developments due to nonconvexity in optimization.

This suggests a hybrid approach of random feature and neural network, called *semi-random feature* (or semi-random unit), to learn representation (or feature) from datasets. The goal is to obtain a new type of basis functions which can retain some theoretical guarantees via injected randomness (or diversity) in hidden weights, while at the same time have the ability to adapt to the data at hand. More concretely, semi-random features are defined as

$$\phi_s(x; \mathbf{r}, \mathbf{w}) = \sigma_s(\mathbf{x}^\top \mathbf{r}) \left(\mathbf{x}^\top \mathbf{w}\right), \tag{1}$$

where $\mathbf{x} = (1, x^\top)^\top$ is assumed to be in $\mathbb{R}^{1+d}$, $\mathbf{r} = (r_0, r^\top)^\top$ is sampled randomly, and $\mathbf{w} = (w_0, w^\top)^\top$ is adjustable weights to be learned from data (hence, it is "semi-random"). Furthermore, the family of functions $\sigma_s$ for $s \in \{0, 1, 2, \dots\}$ is defined as $\sigma_s(z) = (z)^s H(z)$, where $H$ is Heaviside step function ($H(z) = 1$ for $z > 0$ and 0 otherwise). For instance, $\sigma_0$ is simply Heaviside step function, $\sigma_1$ is ramp function, and so on. We call the corresponding semi-random features with $s = 0$ "linear semi-random features (LSR)" and with $s = 1$ "squared semi-random features (SSR)". An illustration of example semi-random features can be found in Figure 4 in Appendix.

Unlike dropout, which uses a data independent random switching mechanism (during training), the random switching in semi-random feature depends on the input data $x$ (during both training and testing), inducing highly nonlinear models with practical advantages. By further enhancing this property, we additionally introduce linear semi-random implicit-ensemble (LSR-IE) features in Section "Image classification benchmarks".

Intuitively, models with semi-random features have more expressive power than those with random features because of the learnable unit parameter $\mathbf{w}$. Yet, these models are less flexible compared to neural networks, since the parameters in $\sigma_s(\mathbf{x}^\top \mathbf{r})$ is sampled randomly. Depending on the problems. this property of semi-random feature *can* result in exponential advantage over fully random feature in *expressiveness* (as discussed in Section "Better than Random Feature?"), and exponential advantage over deep ReLU models in *generalization* error bound (as discussed in Section "Generalization Guarantee").

## One Hidden Layer Model

With semi-random features $\phi_s$ in equation (1), we define one hidden layer model for nonlinear function as

$$\hat{f}_n^s(x; w) = \sum_{k=1}^n \phi_s(x; \mathbf{r}_k, \mathbf{w}_k^{(1)}) w_k^{(2)}, \tag{2}$$

where $\mathbf{r}_k$ is sampled randomly for $k \in \{2, 3, \cdots, n\}$ as described in Section "Semi-Random Features", and $\mathbf{r}_1$ is fixed to be the first element of the standard basis as $\mathbf{r}_1 = \mathbf{e}_1 = (1, 0, 0, \cdots 0)^\top$ (to compactly represent a constant term in $x$). We can think of this model as one hidden layer model by considering $\phi_s(x; \mathbf{r}, \mathbf{w}_k^{(1)})$ as the output of $k$-th unit of the hidden layer, and $\mathbf{w}_k^{(1)}$ as adjustable parameters associated with this hidden layer unit. This way of understanding the model will become helpful when we generalize it to a multilayer model in Section "Multilayer Model". Note that $\hat{f}_n^s(x; w)$ is a nonlinear function of $x$. When it is clear, by the notation $w$, we denote all adjustable parameters in the entire model.

In matrix notation, the model in (2) can be rewritten as

$$\hat{f}_n^s(x; w) = \left(\sigma_s(\mathbf{x}^\top \mathbf{R}) \odot (\mathbf{x}^\top \mathbf{W}^{(1)})\right) W^{(2)}, \qquad (3)$$

where

$$\mathbf{W}^{(1)} = (\mathbf{w}_1^{(1)}, \mathbf{w}_2^{(1)}, \dots, \mathbf{w}_n^{(1)}) \in \mathbb{R}^{(d+1)\times n},$$

$$\mathbf{R} = (\mathbf{r}_1, \mathbf{r}_2, \dots, \mathbf{r}_n) \in \mathbb{R}^{(d+1)\times n}, \text{and}$$

$$\mathbf{W}^{(2)} = (w_1^{(2)}, \dots, w_n^{(2)})^\top \in \mathbb{R}^{n+1}.$$

Here, $(M_1 \odot M_2)$ represents a Hadamard product of two matrices $M_1$ and $M_2$. Furthermore $\sigma_s(M)_{ij} = \sigma_s(M_{ij})$, given a matrix $M$ of any size (with overloads of the symbol $\sigma_s$).

In the following subsections, we present our theoretical results for one hidden layer model. All proofs in this paper are deferred to the appendix.

## Universal Approximation Ability

We show that our model class has universal approximation ability. Given a finite $s$, our model class is defined as

$$\mathcal{F}_n^s = \{x \mapsto \hat{f}_n^s(x; w) \,|\, w \in \mathbb{R}^{d_w}\},$$

where $d_w = (d+1)n + n$ is the number of adjustable parameters. Let $L^2(\Omega)$ be the space of square integrable functions on a compact set $\Omega \subseteq \mathbb{R}^d$. Then Theorem 1 states that we can approximate any $f \in L^2(\Omega)$ arbitrarily well as we increase the number of units $n$. We discuss the importance of the bias term $r_0$ to obtain the universal approximation power in Appendix.

**Theorem 1** (Universal approximation) *Let $s$ be any fixed finite integer and let $\Omega \neq \{0\}$ be any fixed nonempty compact subset of $\mathbb{R}^d$. Then, for any $f \in L^2(\Omega)$, with probability one,*

$$\lim_{n\to\infty} \inf_{\hat{f}\in\mathcal{F}_n^s} \|f - \hat{f}\|_{L^2(\Omega)} = 0.$$

## Optimization Theory

As we have confirmed universal approximation ability of our model class $\mathcal{F}_n^s$ in the previous section, we now want to find a good $\hat{f} \in \mathcal{F}_n^s$ via empirical loss minimization. More specifically, given a dataset $\{(x_i, y_i)\}_{i=1}^m$, we will consider the following optimization problem:

$$\underset{w\in\mathbb{R}^{d_w}}{\text{minimize}}\ \mathcal{L}(w) = \frac{1}{2m}\sum_{i=1}^m \left(y_i - \hat{f}_n^s(x_i; w)\right)^2.$$

Let $Y = (y_1, y_2, \dots, y_m)^\top \in \mathbb{R}^m$ and $\hat{Y} = (f_n^s(x_1; w), f_n^s(x_2; w), \dots, f_n^s(x_m; w))^\top \in \mathbb{R}^m$. Given a matrix $M$, let $\mathbf{P}_{\text{col}(M)}$ and $\mathbf{P}_{\text{null}(M)}$ be the projection matrices onto the column space and null space of $M$.

Our optimization problem turns out to be characterized by the following $m$ by $nd$ matrix:

$$D = \begin{bmatrix} \sigma_s(\mathbf{x}_1^\top \mathbf{r}_1)\mathbf{x}_1^\top & \cdots & \sigma_s(\mathbf{x}_1^\top \mathbf{r}_n)\mathbf{x}_1^\top \\ \vdots & \ddots & \vdots \\ \sigma_s(\mathbf{x}_m^\top \mathbf{r}_1)\mathbf{x}_m^\top & \cdots & \sigma_s(\mathbf{x}_m^\top \mathbf{r}_n)\mathbf{x}_m^\top \end{bmatrix}, \qquad (4)$$

where $\sigma_s(\mathbf{x}_i^\top \mathbf{r}_j)\mathbf{x}_i^\top$ is a $1 \times d$ block at $(i, j)$-th block entry. That is, at any global minimum, we have $\mathcal{L}(w) = \frac{1}{2m}\|\mathbf{P}_{\ker(D^T)} Y\|^2$ and $\hat{Y} = \mathbf{P}_{\text{col}(D)} Y$. Moreover, we can achieve a global minimum in polynomial time in $d_w$ based on the following theorem.

**Theorem 2** (No bad local minima and few bad critical points) *For any $s$ and any $n$, the optimization problem of $\mathcal{L}(w)$ has the following properties:*

*(i) it is non-convex (if $D \neq 0$)*[1]*,*

*(ii) every local minimum is a global minimum,*

*(iii) if $w_k^{(2)} \neq 0$ for all $k \in \{1, 2, \dots, n\}$, every critical point is a global minimum, and*

*(iv) at any global minimum, $\mathcal{L}(w) = \frac{1}{2m}\|\mathbf{P}_{\text{null}(D^T)} Y\|^2$ and $\hat{Y} = \mathbf{P}_{\text{col}(D)} Y$.*

Theorem 2 (optimization) together with Theorem 1 (universality) suggests that not only does our model class contain an arbitrarily good function (as $n$ increases), but also we can find the best function in the model class given a dataset. In the context of understanding the loss surface of neural networks (Kawaguchi 2016; Freeman and Bruna 2016), Theorem 2 implies that the potential problems in the loss surface are due to the inclusion of $\mathbf{r}$ as an optimization variable.

## Generalization Guarantee

In the previous sections, we have shown that our model class has universal approximation ability and that we can learn the best model given a finite dataset. A major remaining question is about the generalization property; how well can a learned model generalize to unseen new observations? Theorem 3 bounds the generalization error; the difference between expected risk, $\frac{1}{2}\mathbb{E}_x(f(x) - \hat{f}(x; w^*))^2$, and empirical risk, $\mathcal{L}(w)$. In Theorem 3, we use the following notations: $[\![\sigma, x]\!] = [\sigma_s(\mathbf{x}^\top \mathbf{r}_1)\mathbf{x} \cdots \sigma_s(\mathbf{x}^\top \mathbf{r}_n)\mathbf{x}] \in \mathbb{R}^{nd}$ and $[\![w]\!] = (w_1^{(2)} w_1^{(1)\top}, w_2^{(2)} w_2^{(1)\top}, \dots, w_n^{(2)} w_n^{(1)\top})^\top \in \mathbb{R}^{nd}$.

**Theorem 3** (Generalization bound for shallow model) *Let $s \geq 0$ and $n > 0$ be fixed. Consider any model class $\mathcal{F}_n^s$ with $\|[\![w]\!]\|_2 \leq C_W$ and $\|[\![\sigma, x]\!]\|_2 \leq C_{\sigma x}$ almost surely. Then, with probability at least $1 - \delta$, for any $\hat{f} \in \mathcal{F}_n^s$,*

$$\frac{1}{2}\mathbb{E}_x(f(x) - \hat{f}(x; w))^2 - \mathcal{L}(w)$$

$$\leq (C_Y^2 + C_{\hat{Y}}^2)\sqrt{\frac{\log\frac{1}{\delta}}{2m}} + 2(C_Y + C_{\hat{Y}})\frac{C_{\hat{Y}}}{\sqrt{m}},$$

*where $C_{\hat{Y}} = C_W C_{\sigma x}$.*

[1]In the case of $D = 0$, $\mathcal{L}(w)$ is convex in a trivial way; our model class only contains a single constant function $x \mapsto 0$.

By combining Theorem 2 (optimization) and Theorem 3 (generalization), we obtain the following remark.

**Remark 1.** *(Expected risk bound)* Let $C_W$ be values such that the global minimal value $\mathcal{L}(w) = \frac{1}{2m}\|\mathbf{P}_{\ker(D^T)}Y\|^2$ is attainable in the model class (e.g., setting $C_W = \|(D^\top D)^\dagger D^\top Y\|$ suffices). Then, at any critical points with $w_k^{(2)} \neq 0$ for all $k \in \{1, 2, \ldots, n\}$ and any local minimum such that $\|[\![w]\!]\|_2 < C_W$, we have

$$\mathbb{E}_x(f(x) - \hat{f}(x; w^*))^2 \leq \frac{\|\mathbf{P}_{\mathrm{null}(D^T)}Y\|^2}{m} + O\left(\sqrt{\frac{\log \frac{1}{\delta}}{m}}\right), \quad (5)$$

with probability at least $1 - \delta$. Here, $O(\cdot)$ notation simply hides the constants in Theorem 3.

In the right-hand side of equation (5), the second term goes to zero as $m$ increases, and the first term goes to zero as $n$ or $d$ increases (because $\mathrm{null}(D^T)$ becomes a smaller and smaller space as $n$ or $d$ increases, eventually containing only 0). Hence, we can minimize the expected risk to zero.

## Multilayer Model

We generalize one hidden layer model to $H$ hidden layer model by composing semi-random features in a nested fashion. More specifically, let $n_l$ be the number of units, or width, in the $l$-th hidden layer for all $l = 1, 2, \ldots, H$. Then we will denote a model of fully-connected feedforward semi-random networks with $H$ hidden layers by

$$\hat{f}^s_{n_1,\ldots,n_H}(x; w) = h_w^{(H)} W^{(H+1)}, \quad (6)$$

where for all $l \in \{2, 3, \cdots, H\}$,

$$h_w^{(l)}(x) = h_r^{(l)}(x) \odot (h_w^{(l-1)}(x)\, W^{(l)}) \text{ and}$$
$$h_r^{(l)}(x) = \sigma_s(h_r^{(l-1)}(x)\, R^{(l)})$$

is the output of the $l$-th *semi-random* hidden layer, and the output of the $l$-th *random* switching layer respectively. Here, $W^{(l)} = (w_1^{(l)}, w_2^{(l)}, \ldots, w_{n_l}^{(l)}) \in \mathbb{R}^{n_{l-1}\times n_l}$ and $R^{(l)} = (r_1^{(l)}, r_2^{(l)}, \ldots, r_{n_l}^{(l)}) \in \mathbb{R}^{n_{l-1}\times n_l}$. Similarly to one hidden layer model, $r_k^{(l)}$ is sampled randomly for $k \in \{2, 3, \ldots, n_l\}$ but $r_1^{(l)}$ is fixed to be $\mathbf{e}_1 = (1, 0, 0, \ldots, 0)^\top$ (to compactly write the effect of constant terms in $x$). The output of the first hidden layer is the same as that of one hidden layer model:

$$h_w^{(1)}(x) = h_r^{(1)}(x) \odot (\mathbf{x}\mathbf{W}^{(1)}) \text{ and } h_r^{(1)}(x) = \sigma_s(\mathbf{x}\mathbf{R}^{(1)}),$$

where the boldface notation emphasizes that we require the bias terms at least in the first layer. In other words, we keep randomly updating the random switching layer $h_r^{(l)}(x)$, and couple it with a linearly adjustable hidden layer $h_w^{(l)}(x)W^{(l)}$ to obtain the next semi-random hidden layer $h_w^{(l+1)}(x)$.

*Convolutional* semi-random feedforward neural networks can be defined in the same way as in equation (6) with vector-matrix multiplication being replaced by $c$ dimensional convolution (for some number $c$). In our experiments, we will test both convolutional semi-random networks as well as fully-connected versions. We will discuss further generalizations of our architecture in Appendix.

### Benefit of Depth

We first confirm that our multilayer model class

$$\mathcal{F}^s_{n_1,\ldots,n_H} = \{x \mapsto \hat{f}^s_{n_1,\ldots,n_H}(x; w) | w \in \mathbb{R}^{d_w}\}$$

preserves universal approximation ability.

**Corollary 4** (Universal approximation with deep model) *Let $s$ be any fixed finite integer and let $\Omega \neq \{0\}$ be any fixed nonempty compact subset of $\mathbb{R}^d$. Then, for any $f \in L^2(\Omega)$, with probability one,*

$$\lim_{n_1,\ldots,n_H\to\infty} \inf_{\hat{f}\in\mathcal{F}^s_{n_1,\ldots,n_H}} \|f - \hat{f}\|_{L^2(\Omega)} = 0.$$

We now know that both of one hidden layer models and deeper models have universal approximation ability. Then, a natural question arises: how can depth benefit us? To answer the question, note that $H$ hidden layer model only needs $O(nH)$ number of parameters (by setting $n = n_1, n_2, \cdots, n_H$) to create around $n^H d$ paths, whereas one hidden layer model requires $O(n^H)$ number of parameters to do so. Intuitively, because of this, the expressive power would grow exponentially in depth $H$, if those exponential paths are not redundant to each other. The redundancy among the paths would be minimized via randomness in the switching and exponentially many combinations of nonlinearities $\sigma_s$.

We formalize this intuition by considering concrete degrees of approximation powers for our models. To do so, we adopt a type of a degree of "smoothness" on the target functions from a previous work (Barron 1993). Consider Fourier representation of a target function as $f(x) = \int_{\omega\in\mathbb{R}^d} \tilde{f}(\omega)e^{i\omega^\top x}$. Define a class of smooth functions $\Gamma_C$:

$$\Gamma_C = \left\{x \mapsto f(x) : \int_{\omega\in\mathbb{R}^d} \|\omega\|_2 |\tilde{f}(\omega)| \leq C\right\}.$$

Any $f \in \Gamma_C$ with finite $C$ is continuously differentiable in $\mathbb{R}^d$, and the gradient of $f$ can be written as $\nabla_x f(x) = \int_{\omega\in\mathbb{R}^d} i\omega\tilde{f}(\omega)e^{i\omega^\top x}$. Thus, via Plancherel theorem, we can view $C$ as the bound on $\|\nabla_x f(x)\|_{L(\mathbb{R}^d)}$. See the previous work (Barron 1993) for a more detailed discussion on the properties of this function class $\Gamma_C$. Theorem 5 states that a lower bound on the approximation power gets better exponentially in depth $H$.

**Theorem 5** (Lower bound on universal approximation power) *Let $\Omega = [0, 1]^d$. For every fixed finite integer $s$, for any depth $H \geq 0$, and for any set of nonzero widths $\{n_1, n_2, \ldots, n_H\}$,*

$$\sup_{f\in\Gamma_C} \inf_{\hat{f}\in\mathcal{F}^s_{n_1\cdots n_H}} \|f - \hat{f}\|_{L^2(\Omega)} \geq \frac{\kappa C}{d^2}\left(\prod_{l=1}^{H} n_l\right)^{-1/d},$$

*where $\kappa \geq (8\pi e^{(\pi-1)})^{-1}$ is a constant.*

By setting $n = n_1 = \cdots = n_H$, the lower bound in Theorem 5 becomes: $\sup_{f\in\Gamma_C} \inf_{\hat{f}\in\mathcal{F}^s_{n_1\cdots n_H}} \|f - \hat{f}\|_{L^2(\Omega)} \geq \frac{\kappa C}{d^2} n^{-H/d}$, where we can easily see the benefit of the depth $H$.

However, this analysis is still preliminary for the purpose of fully understanding the benefit of the depth. Our hope here is to provide a formal statement to aid intuitions. To help our intuitions further, we discuss about an upper bound on universal approximation power for multilayer model in Appendix.

## Optimization Theory

Similarly to one hidden layer case, we consider the following optimization problem:

$$\underset{w\in\mathbb{R}^{d_w}}{\text{minimize}}\, \mathcal{L}^{(H)}(w) = \frac{1}{2m}\sum_{i=1}^{m}\left(y_i - f^{s}_{n_1,\ldots,n_H}(x;w)\right)^2.$$

Compared to one hidden layer case, our theoretical understanding of multilayer model is rather preliminary. Here, given a function $g(w_1, w_2, \ldots, w_n)$, we say that $\bar{w} = (\bar{w}_1, \bar{w}_2, \ldots, \bar{w}_n)$ is a global minimum of $g$ with respect to $w_1$ if $\bar{w}_1$ is a global minimum of $\tilde{g}(w_1) = g(w_1, \bar{w}_2, \ldots, \bar{w}_n)$.

**Corollary 6** (No bad local minima and few bad critical points w.r.t. last two layers) *For any $s$, any depth $H \geq 1$, and any set of nonzero widths $\{n_1, n_2, \ldots, n_H\}$, the optimization problem of $\mathcal{L}^{(H)}(w)$ has the following property:*

*(i) every local minimum is a global minimum with respect to $(W^{(H)}, W^{(H+1)})$, and*

*(ii) if $w_k^{(H+1)} \neq 0$ for all $k \in \{1, 2, \ldots, n_H\}$, every critical point is a global minimum with respect to $(W^{(H)}, W^{(H+1)})$.*

Future work is still required to investigate the theoretical nature of the optimization problem with respect to all parameters. Some hardness results of a standard neural network optimization come from the difficulty of learning activation pattern via optimization of the variable $\mathbf{r}$ (Livni, Shalev-Shwartz, and Shamir 2014). In this sense, our optimization problem is somewhat easier, and it would be interesting to see if we can establish meaningful optimization theory for semi-random model as a first step to establish the theory for neural networks in general.

## Generalization Guarantee

The following corollary bounds the generalization error. In the statement of the corollary, we can easily see that the generalization error goes to zero as $m$ increases (as long as the relevant norms are bounded). Hence, we can achieve generalization. In Corollary 7, we use the following notations: $[\![\sigma, x]\!]_{k_0,k_1,\ldots,k_H} = [\![\sigma]\!]_{k_1,\ldots,k_H}(x)\mathbf{x}_{k_0}$ and $[\![w]\!]_{k_0,k_1,\ldots,k_H} = \left(\prod_{l=1}^{H} w^{(l)}_{k_{l-1}k_l}\right) w^{(H+1)}_{k_H}$. Let $\mathrm{vec}(M)$ be a vectorization of a tensor $M$.

**Corollary 7** (Generalization bound for deep model) *Let $s \geq 0$ and $H \geq 1$ be fixed. Let $n_l > 0$ be fixed for $l = 1, 2, \ldots, H$. Consider the model class $\mathcal{F}^{s}_{n_1,\ldots,n_H}$ with $\|\mathrm{vec}([\![w]\!])\|_2 \leq C_W$ and $\|\mathrm{vec}([\![\sigma, x]\!])\|_2 \leq C_{\sigma x}$ almost surely. Then, with probability at least $1-\delta$, for any $\hat{f} \in \mathcal{F}^{s}_{n_1,\ldots,n_H}$,*

$$\frac{1}{2}\mathbb{E}_x(f(x) - \hat{f}(x;w^*))^2 - \mathcal{L}^{(H)}(w) \leq (C_Y^2 + C_{\hat{Y}}^2)\sqrt{\frac{\log\frac{1}{\delta}}{2m}} + 2(C_Y + C_{\hat{Y}})\frac{C_{\hat{Y}}}{\sqrt{m}},$$

*where $C_{\hat{Y}} = C_W C_{\sigma x}$.*

The generalization bound in Corollary 7 can be exponentially better than the known bounds of ReLU; e.g., see (Sun et al. 2016; Neyshabur, Tomioka, and Srebro 2015; Xie, Deng, and Xing 2015). The known generalization upper bounds of ReLU explicitly contain $2^H$ factor, which comes from ReLU nonlinearity and grows exponentially in depth $H$. In contrast, the generalization bound in Corollary 7 does not explicitly depend on any of the depth, the number of trainable weights, and the dimensionality of the domain. Note that $f_{n_1,\ldots,n_H}(x) = vec([\![w]\!])^\top vec([\![\sigma, x]\!]) = \|vec([\![w]\!])\|_2 \|vec([\![\sigma, x]\!])\|_2 \cos\theta$. Hence, even though the dimensionality of $\mathrm{vec}([\![w]\!])$ and $\mathrm{vec}([\![\sigma, x]\!])$ grows exponentially in the depth, the norm bound $C_W C_{\sigma x}$ would stay near the norm of the output. Indeed, $C_W C_{\sigma x} \approx f_{n_1,\ldots,n_H}(x)/\cos\theta$. As a new related work, the generalization bounds of *standard nets trained by a novel two-phase training procedure* in a recent paper (Kawaguchi, Kaelbling, and Bengio 2017) have this desired qualitative property similarly to our generalization bounds of *semi-random nets*.

# Experiments

We compare semi-random features with random features (RF) and neural networks with ReLU on both UCI datasets and image classification benchmarks. We will study two variants of semi-random features for $s = 0$ (LSR: linear semi-random features) and $s = 1$ (SSR: squared semi-random features) in $\sigma_s(\cdot)$ from equation (1). Additional experimental details are presented in Appendix. The source code of the proposed method is publicly available at: http://github.com/zixu1986/semi-random.

## Simple Test Function

We first tested the methods with a simple sine function, $f(x) = \sin(x)$, where we can easily understand what is happening. Figure 1 shows the test errors with one standard deviations. As we can see, semi-random network (LSR) performed the best. The problem of ReLU became clear once we visualized the function learned at each iteration: ReLU network had a difficulty to diversify activation units to mimic the frequent oscillations of the sine function (i.e., it took long time to diversely allocate the breaking points of its piecewise linear function). The visualizations of learned functions at each iteration for each method are presented in Appendix. On average, they took 54.39 (ReLU), 43.04 (random), 45.44 (semi-random) seconds. Their training errors are presented in Appendix.

## UCI datasets

We have comparisons on six large UCI datasets.The network architecture used on this dataset is multi-layer networks with $l = [1, 2, 4]$ hidden layers and $k = [1, 2, 4, 8, 16] \times d$ hidden units per layer where $d$ is the input data dimension.

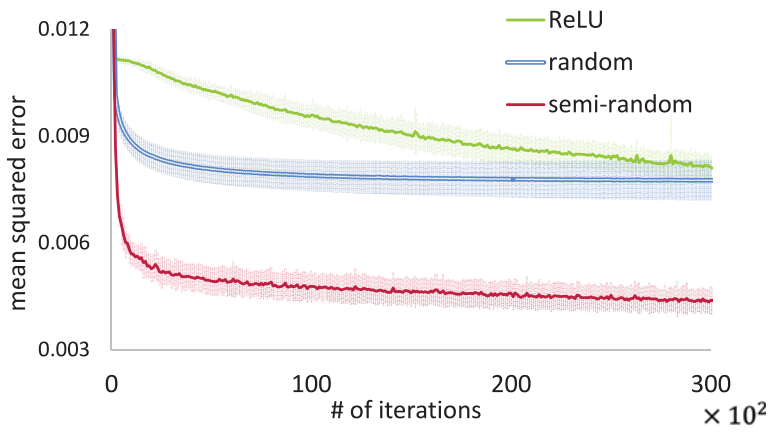

Figure 1: Test error for a simple test function.

Table 1: Performance comparison on UCI datasets. RF for random features, LSR and SSR for linear ($s = 0$) and squared ($s = 1$) semi-random features respectively. $m_{\text{tr}}$: number of training data points; $m_{\text{te}}$: number of test data points; $d$ dimension of the data.

| Dataset | method | err (%) | Dataset | method | err (%) |
|---|---|---|---|---|---|
| covtype | ReLU | 5.6 | adult | ReLU | 15.0 |
| $m_{\text{tr}} = 522,910$ | RF | 20.2 | $m_{\text{tr}} = 32,561$ | RF | 14.9 |
| $m_{\text{te}} = 58,102$ | LSR | 5.7 | $m_{\text{te}} = 16,281$ | LSR | 14.8 |
| $d = 54$ | SSR | 14.4 | $d = 123$ | SSR | 14.9 |
| webspam | ReLU | 1.1 | senseit | ReLU | 13.6 |
| $m_{\text{tr}} = 280,000$ | RF | 6.0 | $m_{\text{tr}} = 78,823$ | RF | 16.0 |
| $m_{\text{te}} = 70,000$ | LSR | 1.1 | $m_{\text{te}} = 19,705$ | LSR | 13.9 |
| $d = 123$ | SSR | 2.2 | $d = 100$ | SSR | 13.3 |
| letter | ReLU | 13.1 | sensor | ReLU | 2.0 |
| $m_{\text{tr}} = 15,000$ | RF | 14.9 | $m_{\text{tr}} = 48,509$ | RF | 13.4 |
| $m_{\text{tr}} = 5,000$ | LSR | 6.5 | $m_{\text{tr}} = 10,000$ | LSR | 1.4 |
| $d = 16$ | SSR | 5.6 | $d = 48$ | SSR | 5.7 |

**Comparison of best performance.** In Table 1, we listed the best performance among different architectures for semi-random and fully random methods. For ReLU, we fix the number of hidden units to be $d$. On most datasets, semi-random features achieve smaller errors compared with ReLU by using more units. In addition, semi-random units have significant lower errors than random features.

**Matching the performance of ReLU.** The top row of Figure 2 demonstrates how many more units are required for random and semi-random features to reach the test errors of networks with ReLU. First, all three methods enjoy lower test errors by increasing the number of hidden units. Second, semi-random units can achieve comparable performance to ReLU with slightly more units, around 2 to 4 times in Webspam dataset. In comparison, random features require many more units, more than 16 times. These experiments clearly show the benefit of adaptivity in semi-random features.

**Depth vs width.** The bottom row of Figure 2 explores the benefit of depth. Here, "$l$-layer" indicates $l$ *hidden* layer model. To grow the number of total units, we can either use more layers or more units per layer. Experiment results suggest that we can gain more in performance by going deeper. The ability to benefit from deeper architecture is an important feature that is not possessed by random features. The details on how the test error changes w.r.t. the number of layers and number of units per layer are shown in Figure 3. As we can see, on most datasets, more layers and more units lead to smaller test errors. However, the adult dataset is more noisy and it is easier to overfit. All types of neurons perform relatively the same on this dataset, and more parameters actually lead to worse results. Furthermore, the squared semi-random features have very similar error pattern to neural network with ReLU.

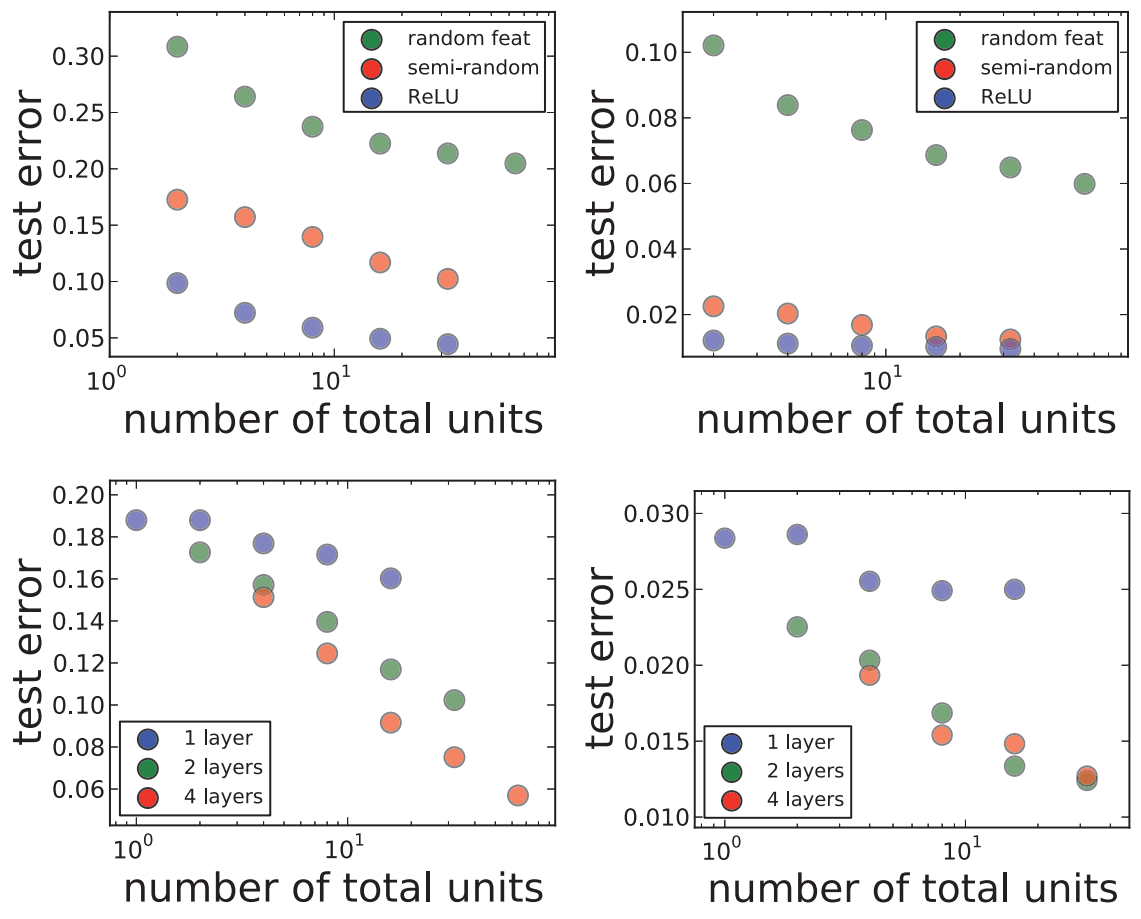

(a) Covtype dataset (b) Webspam dataset

Figure 2: Top row: Linear semi-random features match the performance of ReLU for two hidden layer networks in two datasets. Bottom row: Depth vs width of linear semi-random features. Both plots show performance of semi-random units. Even the total number of units is the same, deeper models achieve lower test error.

Table 2: Test error (in %) of different methods on three image classification benchmark datasets. $2\times$, $4\times$ and $16\times$ mean the number of units used is 2 times, 4 times and 16 times of that used in neural network with ReLU respectively.

| neuron type | MNIST | CIFAR10 | SVHN |
|---|---|---|---|
| ReLU | 0.70 | 16.3 | 3.9 |
| RF | 8.80 | 59.2 | 73.9 |
| RF $2\times$ | 5.71 | 55.8 | 70.5 |
| RF $4\times$ | 4.10 | 49.8 | 58.4 |
| RF $16\times$ | 2.69 | 40.7 | 37.1 |
| LSR | 0.97 | 21.4 | 7.6 |
| LSR $2\times$ | 0.78 | 17.4 | 6.9 |
| LSR $4\times$ | 0.71 | 18.7 | 6.4 |
| LSR-IE | 0.59 | 20.0 | 6.9 |
| LSR-IE $2\times$ | 0.47 | 16.8 | 5.9 |
| LSR-IE $4\times$ | 0.54 | 14.9 | 4.8 |

## Image classification benchmarks

We have also compared different methods on three image classification benchmark datasets. Here we use publicly available and well-tuned neural network architectures from tensorflow for the experiments. We simply replace ReLU by random and semi-random units respectively. The results are summarized in Table 2.

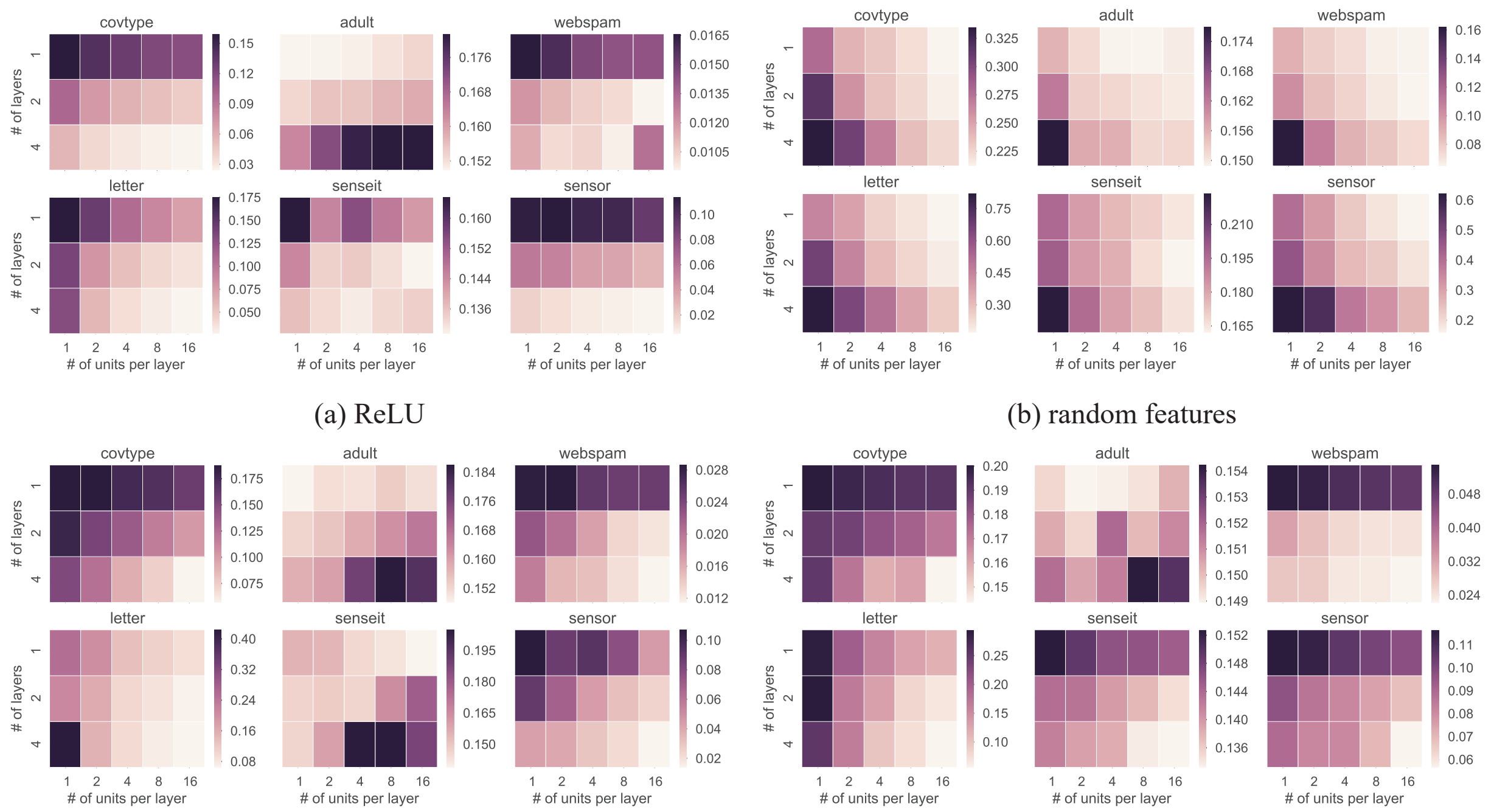

(a) ReLU

(b) random features

(c) linear semi-random features

(d) squared semi-random features

Figure 3: Detailed experiment results for all types of neurons and on all datasets. The heat map for each dataset shows how the test error changes w.r.t. the number of layers and number of units per layer.

**LSR-IE Unit.** To improve the practical performance of LSR unit while preserving its theoretical properties, we additionally introduced a new unit, called linear semi-random *implicit*-ensemble (LSR-IE) unit. Unlike an *explicit* ensemble, LSR-IE trains only a single network and comparable to a single standard network with dropout.

Recall that in LSR unit, we have a single random net (with random weights $\mathbf{r}$) and a single trainable net (with trainable weights $\mathbf{w}$). In contrast, in LSR-IE unit, we increase the number of the random nets (with random weights $\mathbf{r}_k$ for $k = 1, \ldots, K$) *while keeping the number of the trainable net to be one*. That is, by modifying equation (1) of semi-random unit, we implement the LSR-IE unit as follows: for each $k \in \{1, \ldots, K\}$, we have

$$\phi_s(x; \mathbf{r}_k, \mathbf{w}) = \sigma_s(\mathbf{x}^\top \mathbf{r}_k) \left(\mathbf{x}^\top \mathbf{w}\right), \tag{7}$$

and we randomly sample the index $k$ at each step of SGD during training. Then, during test, we average the pre-activation of output units over $k \in \{1, \ldots, K\}$. We set the number of random nets $K$ to be 2 in MNIST and CIFAR-10 and 4 in SVHN.

**MNIST dataset.** MNIST is a popular dataset for recognizing handwritten digits. It contains 28 $\times$ 28 grey images, 60,000 for training and 10,000 for test. We use a convolution neural network consisting of two convolution layers, with 5 $\times$ 5 filters and the number of channels is 32 and 64, respectively. Each convolution is followed by a max-pooling layer, then finally a fully-connected layer of 512 units with 0.5 dropout. Increasing the number of units for semi-random leads to better performance. At four times the size of the original network, semi-random feature can achieve very close errors of $0.71\%$. In contrast, even when increasing the number of units to 16 times more, random features still cannot reach below $1\%$.

**CIFAR10 dataset.** CIFAR 10 contains internet images and consists of 50,000 32 $\times$ 32 color images for training and 10,000 images for test. We use a convolutional neural network architecture with two convolution layers, each with 64 5 $\times$ 5 filters and followed by max-pooling. The fully-connected layers contain 384 and 192 units. By using two times more units, semi-random features are able to achieve similar performance with ReLU. However, the performance of random features lags behind by a huge margin.

**SVHN dataset.** The Street View House Numbers (SVHN) dataset contains house digits collected by Google Street View. We use the $32 \times 32$ color images version and only predict the digits in the middle of the image. For training, we combined the training set and the extra set to get a dataset with 604,388 images. We use the same architecture as in the CIFAR10 experiments.

## Conclusion

In this paper, we proposed the method of semi-random features. For one hidden layer model, we proved that our model class contains an arbitrarily good function as the width increases (universality), and we can find such a good function (optimization theory) that generalizes to unseen new data (generalization bound). For deep model, we proved universal approximation ability, a lower bound on approximation error, a partial optimization guarantee, and a generalization bound. Furthermore, we demonstrated the advantage of semi-random features over fully-random features via empirical results and theoretical insights.

The idea of semi-random feature itself is more general than what is explored in this paper, and it opens up several intersecting directions for future work. Indeed, we can generalize any deep architecture by having an option to include *semi-random units* per unit level. We can also define a more general semi-random feature as: given some nonconstant functions $\sigma$ and $g$,

$$\phi(x; \mathbf{r}, \mathbf{w}) = \sigma(\mathbf{x}^\top \mathbf{r}) g\left(\mathbf{x}^\top \mathbf{w}\right),$$

where $\mathbf{x} = (1, x)$ is assumed to be in $\mathbb{R}^{1+d}$, $\mathbf{r} = (r_0, r)$ is sampled randomly, and $\mathbf{w} = (w_0, w)$ is adjustable weights to be learned from data. This general formulation would lead to a flexibility to balance expressivity, generalization and theoretical tractability.

## References

Barron, A. R. 1993. Universal approximation bounds for superpositions of a sigmoidal function. *IEEE Transactions on Information theory* 39(3):930–945.

Bellemare, M. G.; Naddaf, Y.; Veness, J.; and Bowling, M. 2013. The arcade learning environment: An evaluation platform for general agents. *Journal of Artificial Intelligence Research* 47:253–279.

Choromanska, A.; LeCun, Y.; and Arous, G. B. 2015. Open problem: The landscape of the loss surfaces of multilayer networks. In *Proceedings of The 28th Conference on Learning Theory*, 1756–1760.

Deng, J.; Dong, W.; Socher, R.; Li, L.-J.; Li, K.; and Fei-Fei, L. 2009. Imagenet: A large-scale hierarchical image database. In *Computer Vision and Pattern Recognition, 2009*, 248–255. IEEE.

Freeman, C. D., and Bruna, J. 2016. Topology and geometry of half-rectified network optimization. *arXiv preprint arXiv:1611.01540*.

Huang, G.-B.; Chen, L.; Siew, C. K.; et al. 2006. Universal approximation using incremental constructive feedforward networks with random hidden nodes. *IEEE Trans. Neural Networks* 17(4):879–892.

Kawaguchi, K.; Kaelbling, L. P.; and Bengio, Y. 2017. Generalization in deep learning. *arXiv preprint arXiv:1710.05468*.

Kawaguchi, K. 2016. Deep learning without poor local minima. In *Advances in Neural Information Processing Systems*.

LeCun, Y.; Bengio, Y.; and Hinton, G. 2015. Deep learning. *Nature* 521(7553):436–444.

Leshno, M.; Lin, V. Y.; Pinkus, A.; and Schocken, S. 1993. Multilayer feedforward networks with a nonpolynomial activation function can approximate any function. *Neural networks* 6(6):861–867.

Livni, R.; Shalev-Shwartz, S.; and Shamir, O. 2014. On the computational efficiency of training neural networks. In *Advances in Neural Information Processing Systems*, 855–863.

Mohri, M.; Rostamizadeh, A.; and Talwalkar, A. 2012. *Foundations of machine learning*. MIT press.

Neyshabur, B.; Tomioka, R.; and Srebro, N. 2015. Norm-based capacity control in neural networks. In *Proceedings of The 28th Conference on Learning Theory*, 1376–1401.

Pennington, J.; Yu, F.; and Kumar, S. 2015. Spherical random features for polynomial kernels. In *Advances in Neural Information Processing Systems*, 1846–1854.

Rahimi, A., and Recht, B. 2008. Random features for large-scale kernel machines. In *Advances in Neural Information Processing Systems*, 1177–1184.

Rahimi, A., and Recht, B. 2009. Weighted sums of random kitchen sinks: Replacing minimization with randomization in learning. In *Advances in neural information processing systems*, 1313–1320.

Scholkopf, B., and Smola, A. J. 2001. *Learning with kernels: support vector machines, regularization, optimization, and beyond*. MIT press.

Shamir, O. 2016. Distribution-specific hardness of learning neural networks. *arXiv preprint arXiv:1609.01037*.

Shawe-Taylor, J., and Cristianini, N. 2004. *Kernel methods for pattern analysis*. Cambridge university press.

Silver, D.; Huang, A.; Maddison, C. J.; Guez, A.; Sifre, L.; Van Den Driessche, G.; Schrittwieser, J.; Antonoglou, I.; Panneershelvam, V.; Lanctot, M.; et al. 2016. Mastering the game of go with deep neural networks and tree search. *Nature* 529(7587):484–489.

Sindhwani, V.; Avron, H.; and Mahoney, M. W. 2014. Quasi-monte carlo feature maps for shift-invariant kernels. In *International Conference on Machine Learning*.

Sun, S.; Chen, W.; Wang, L.; Liu, X.; and Liu, T.-Y. 2016. On the depth of deep neural networks: a theoretical view. In *Proceedings of the Thirtieth AAAI Conference on Artificial Intelligence*, 2066–2072. AAAI Press.

Swirszcz, G.; Czarnecki, W. M.; and Pascanu, R. 2016. Local minima in training of deep networks. *arXiv preprint arXiv:1611.06310*.

Telgarsky, M. 2016. Benefits of depth in neural networks. In *29th Annual Conference on Learning Theory*, 1517–1539.

Xie, P.; Deng, Y.; and Xing, E. 2015. On the generalization error bounds of neural networks under diversity-inducing mutual angular regularization. *arXiv preprint arXiv:1511.07110*.

Xie, B.; Liang, Y.; and Song, L. 2016. Diversity leads to generalization in neural networks. *arXiv preprint arXiv:1611.03131*.

# Appendix

In practice, we can typically assume that $x \in \Omega \subseteq \mathbb{R}^d$ with some sufficiently large compact subspace $\Omega$. Thus, given such a $\Omega$, we use the following sampling procedure in order for our method to be efficiently executable in practice: $r$ is sampled uniformly from a $d$ dimensional unit sphere $\mathbb{S}^{d-1}$, and $r_0$ is sampled such that the probability measure on any open ball in $\mathbb{R}$ with its center in $[-\text{radius}(\Omega), \text{radius}(\Omega)]$ is nonzero.[2] For example, uniform distribution that covers $[-\text{radius}(\Omega), \text{radius}(\Omega)]$ or normal distribution with a nonzero finite variance suffices the above requirement.

## Visualization of semi-random features

We visualize the surfaces of two semi-random features (a) linear (LSR) ($s = 0$) and (b) square (SSR) ($s = 1$) in Figure 4. The semi-random features are defined by two parts. The random weights and nonlinear thresholding define a hyperplane, where on one side, it is zero and on the other side, it is either an adjustable linear or adjustable square function. During learning, gradient descent is used to tune the adjustable parts to fit target functions.

## Importance of Bias Terms in First Layer

We note the importance of the bias term $r_0$ in the first layer, to obtain universal approximation power. Without the bias term, Theorem 1 does not hold. Indeed, it is easy to see that without the bias term, Heaviside step functions do not form a function class with universal approximation ability for $x \in \mathbb{R}^d$. The importance of the bias term can also be seen by the binomial theorem. For example, let $g(z)$ ($z \in \mathbb{R}$) be a polynomial of degree *at least* $c$ for some constant $c$, then $g$ does *not* form a function class that contains the polynomial of degree less than $c$ in $z$. However, if $z = x + r_0$, it is possible to contain polynomials of degrees less than $c$ in $x$ by binomial theorem.

## Proofs for One Hidden Layer Model

In this section, we provide the proofs of the theorems presented in Section "One Hidden Layer Model". We denote a constant function $x \mapsto 1$ by $\mathbb{1}$ (i.e., $\mathbb{1}(x) = 1$).

### Proof of Theorem 1 (Universal Approximation)

To prove the theorem, we recall the following known result.

**Lemma 8** (Leshno et al. 1993, Proposition 1) *For any $\sigma : \mathbb{R} \to \mathbb{R}$, $\text{span}(\{\sigma(r^\top x + r_0) : r \in \mathbb{R}^d, r_0 \in \mathbb{R}\})$ is dense in $L^p(\Omega)$ for every compact subset $\Omega \subseteq \mathbb{R}^d$ and for every $p \in [1, \infty)$, if and only if $\sigma$ is not a polynomial of finite degree (almost everywhere).*

With this lemma, we first prove that for every fixed $s$, the span of a set of our functions $\sigma_s$ is dense in $L^2(\Omega)$.

[2] The radius of a compact subspace $\Omega$ of $\mathbb{R}^d$ is defined as $\text{radius}(\Omega) = \sup_{x\in\Omega} \|x\|_2$ and a open ball is defined as $B(\bar{z}, \delta) = \{z : \|z - \bar{z}\| < \delta\}$ where $\bar{z}$ is the center of the ball.

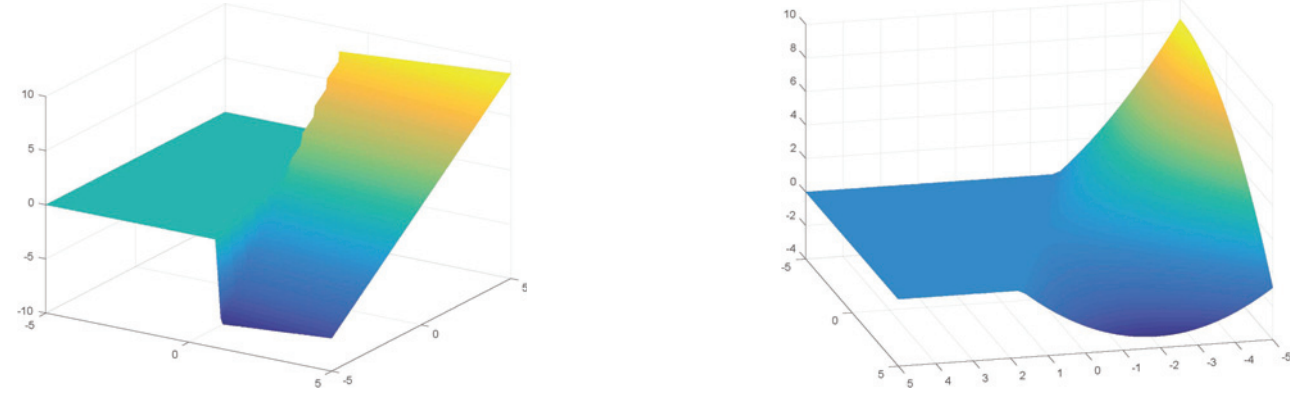

(a) linear semi-random (b) square semi-random

Figure 4: Visualization of linear semi-random (LSR) ($s = 0$) and squared semi-random (SSR) ($s = 1$) units. The random weights define a hyperplane: on one side, it is zero while on the other side, it is either adjustable linear or adjustable square function.

**Lemma 9** *For every fixed finite integer $s$ and for every compact subset $\Omega \subseteq \mathbb{R}^d$, $\text{span}\{\sigma_s(r^\top x + r_0), \mathbb{1}(x) : r \in \mathbb{S}^{d-1}, r_0 \in \mathbb{R}\}$ is dense in $L^2(\Omega)$, where $g(x) = 1$ represents a constant function.*

**Proof** Since $\sigma_0$ is Heaviside step function, $\sigma_0$ is not a polynomial of finite degree and $\text{span}\{\sigma_0(r^\top x + r_0) : r \in \mathbb{R}^d, r_0 \in \mathbb{R}\}$ is dense in $L^2(\Omega)$ (we know each of both statements independently without Lemma 8). For the case of $s \geq 1$, $\sigma_s(z) = z^s \sigma_0(z)$ is not a polynomial of finite degree (i.e., otherwise, $\sigma_0$ is a polynomial of finite degree). Thus, by Lemma 8, for fixed finite integer $s \geq 1$, $\text{span}\{\sigma_s(r^\top x + r_0) : r \in \mathbb{R}^d, r_0 \in \mathbb{R}\}$ is dense in $L^2(\Omega)$.

Since $\sigma_s(\frac{r^\top x}{\|r\|} + r_0) = \|r\|^{-s}\sigma_s(r^\top x + \|r\| r_0)$ for all $r \neq 0$ (because $H(z)$ is insensitive to a positive scaling of $z$),

$$\begin{aligned}&\text{span}\{\sigma_s(r^\top x + r_0) : r \in \mathbb{R}^d, r_0 \in \mathbb{R}\}\\&= \text{span}\{\sigma_s(r^\top x + r_0) : r \in \mathbb{S}^{d-1} \cup \{0\}, r_0 \in \mathbb{R}\}\\&= \text{span}\{\sigma_s(r^\top x + r_0), \mathbb{1}(x) : r \in \mathbb{S}^{d-1}, r_0 \in \mathbb{R}\}.\end{aligned}$$

This completes the proof. ■

In practice, we want to avoid wasting samples; we want to have a choice to sample $r_0$ from the interval that matters. In order to do that, we admit the dependence of our function class on $\Omega$ via the following lemma.

**Lemma 10** *For every fixed finite integer $s$ and for every compact subset $\Omega \subseteq \mathbb{R}^d$, $\text{span}\{\sigma_s(r^\top x + r_0), \mathbb{1}(x) : r \in \mathbb{S}^{d-1}, |r_0| \leq \text{radius}(\Omega)\}$ is dense in $L^2(\Omega)$.*

**Proof** For any $r_0 \in \mathbb{R} \setminus [-\text{radius}(\Omega), +\text{radius}(\Omega)]$, since $|r^\top x| \leq \|r\|_2 \|x\|_2 \leq \text{radius}(\Omega)$, either $\sigma_s(r^\top x + r_0) = 0$ or $= (r^\top x + r_0)^s$, for all $x \in \Omega$. Meanwhile, with $r_0 = \text{radius}(\Omega)$, $\sigma_s(r^\top x + r_0) = (r^\top x + r_0)^s$, for all $x \in \Omega$. Since $(r^\top x + r_0)^s = \sum_{k=0}^{s} \binom{s}{k} r_0^{s-k} (r^\top x)^k$, it is a polynomial in $x$, and $r_0$ only affects the coefficients of the polynomial. Thus, for any $x \in \Omega$, $\text{span}\{\sigma_s(r^\top x + r_0), \mathbb{1}(x) : r \in \mathbb{S}^{d-1}, r_0 \in \mathbb{R}\} = \text{span}\{\sigma_s(r^\top x + r_0), \mathbb{1}(x) : r \in \mathbb{S}^{d-1}, |r_0| \leq \text{radius}(\Omega)\}$, because the difference in the nonzero coefficients of the polynomials can be taken care of by adjusting the coefficients of the linear combination, unless the right-hand side

only contains $r_0 \in \{0\}$. Having only $r_0 \in \{0\}$ implies $\Omega = \{0\}$, which is assumed to be false. ■

We use the following lemma to translate the statement with uncountable samples in Lemma 10 to countable samples in Theorem 1.

**Lemma 11** *Let $s$ be a fixed finite integer, $\Omega$ be any compact subset of $\mathbb{R}^d$, and $(\bar{r}, \bar{r}_0)$ is in*

$$\mathbb{S}^{d-1} \times [-\text{radius}(\Omega), \text{radius}(\Omega)].$$

*Then, for any $\epsilon > 0$, there exists a $\delta$ such that for any $(r, r_0) \in B((\bar{r}, \bar{r}_0), \delta)$,*

$$\int_{x \in \Omega} (\sigma_s(r^\top x + r_0) - \sigma_s(\bar{r}^\top x + \bar{r}_0))^2 < \epsilon.$$

**Proof** For the case of $s = 0$, it directly follows the proof of lemma II.5 in (Huang et al. 2006). For the case of $s \geq 1$, we can think of $\sigma_s$ as a uniformly continuous function with a compact domain that contains all the possible inputs of $\sigma_s$ for our choice of $\Omega$ and $\mathbb{S}^{d-1} \times [-\text{radius}(\Omega), \text{radius}(\Omega)]$. The lemma immediately follows from its uniform continuity. ■

Note that if $s \to \infty$, $\|x\| \to \infty$, $\|r\| \to \infty$, or $\|r_0\| \to \infty$, the proof of Lemma 11 does not work. We are now ready to prove Theorem 1.

**Proof of Theorem 1** Fix $s$ and let $\sigma_i : x \mapsto \sigma_s(r_i^\top x + r_{0i})$ with $r_i$ and $r_{0i}$ being sampled randomly for $i = 2, 3, \ldots$ as specified in Section "Semi-Random Features". Let $\sigma_1 : x \mapsto \sigma_s(r_1^\top x + r_{01}) = \mathbb{1}(x) = 1$ as specified in Section "One Hidden Layer Model". Since $\text{span}\{\sigma_1, \sigma_2, ..., \sigma_n\} \subseteq \mathcal{F}_n^s$ (as we can get $\text{span}\{\sigma_1, \sigma_2, ..., \sigma_n\}$ by only using $w_{0i}^{(1)}$ and $w_i^{(2)}$), we only need to prove the universality of $\text{span}\{\sigma_1, \sigma_2, ..., \sigma_n\}$ as $n \to \infty$. We prove the statement by contradiction. Suppose that $\text{span}\{\sigma_1, \sigma_2, ...\}$ is not dense in $L^2(\Omega)$ (with some nonzero probability). Then, there exists a nonzero $f_0 \in L^2(\Omega)$ such that $\langle f_0, \sigma_i \rangle = 0$ for all $i = 1, 2, \ldots$. However, from Lemma 10, either $\langle f_0, \sigma_1 \rangle \neq 0$ or there exists $(\bar{r}, \bar{r}_0) \in \mathbb{S}^{d-1} \times [-\text{radius}(\Omega), \text{radius}(\Omega)]$ such that $|\langle f_0, \sigma_{\bar{r}, \bar{r}_0} \rangle| = |c| > 0$ with some constant $c$, where $\sigma_{\bar{r}, \bar{r}_0} = \sigma_s(\bar{r}^\top x + \bar{r}_0)$. If $\langle f_0, \sigma_1 \rangle \neq 0$, we get the desired contradiction, and hence we considerer the latter case. In the latter case, for any $\epsilon > 0$,

$$\begin{aligned}
|c| &= |\langle f_0, \sigma_{\bar{r}, \bar{r}_0} \rangle| \\
&= |\langle f_0, \sigma_{\bar{r}, \bar{r}_0} \rangle - \langle f_0, \sigma_i \rangle| \ \ \forall i \in \{2, 3, ...\} \\
&= |\langle f_0, \sigma_{\bar{r}, \bar{r}_0} - \sigma_i \rangle| \ \ \forall i \in \{2, 3, ...\} \\
&\leq \|\sigma_{\bar{r}, \bar{r}_0} - \sigma_i\| \|f_0\| \ \ \forall i \in \{2, 3, ...\} \\
&< \epsilon \|f_0\| \ \ \exists i \in \{2, 3, ...\},
\end{aligned}$$

where the last line follows from lemma 11 and our sampling procedure; that is, for any $\epsilon > 0$, if a sampled $(r_i, r_{0i})$ is in a $\delta$-ball, $\|\sigma_i - \sigma_{\bar{r}, \bar{r}_0}\| < \epsilon$ (from Lemma 11). Since our sampling procedure allocates nonzero probability on any such interval, as $i \to \infty$, the probability of sampling $(r_i, r_{0i})$ in any $\delta$-ball becomes one. This shows the last line. The above inequality leads a contradiction $|c| < |c|$ by choosing $\epsilon < |c| / \|f_0\|$. This completes the proof. ■

## Proof of Theorem 2 (No Bad Local Minima and Few Bad Critical Points)

Since multiplying a constant in $w$ does not change the optimization problem (in terms of optimizer), in the proof of Theorem 2, we consider

$$\mathcal{L}(w) := \sum_{i=1}^{m} \left( y_i - \hat{f}_n^s(x_i; w) \right)^2,$$

to be succinct.

**Proof of Theorem 2 (iii) and (iv)** Define a $m \times (nd)$ matrix as

$$D' = \begin{bmatrix} w_1^{(2)} \sigma_s(\mathbf{x}_1^\top \mathbf{r}_1) \mathbf{x}_1^\top & \cdots & w_n^{(2)} \sigma_s(\mathbf{x}_1^\top \mathbf{r}_n) \mathbf{x}_1^\top \\ \vdots & \ddots & \vdots \\ w_1^{(2)} \sigma_s(\mathbf{x}_m^\top \mathbf{r}_1) \mathbf{x}_m^\top & \cdots & w_n^{(2)} \sigma_s(\mathbf{x}_m^\top \mathbf{r}_n) \mathbf{x}_m^\top \end{bmatrix}.$$

where $w_j^{(2)} \sigma_s(\mathbf{x}_i^\top \mathbf{r}_j) \mathbf{x}_i^\top$ is a $1 \times d$ block at $(i, j)$-th block entry. Then, we can rewrite the objective function as

$$\mathcal{L}(w) = \|D' \text{vec}(w^{(1)}) - Y\|_2^2,$$

where

$$\text{vec}(w^{(1)}) = (w_1^{(1)\top}, w_2^{(1)\top}, \ldots, w_n^{(1)\top})^\top.$$

Therefore, at any critical point w.r.t. $w^{(1)}$ (i.e., by taking gradient of $\mathcal{L}$ w.r.t. $w^{(1)}$ and setting it to zero), $\hat{Y}$ is the projection of $Y$ onto the column space of $D'$.

Since $\hat{f}_n(x_i; w) = \sum_{k=1}^n \sigma_s(\mathbf{x}_i^\top \mathbf{r}_k)(\mathbf{x}_i^\top \mathbf{w}_k^{(1)}) w_k^{(2)}$,

$$\mathcal{L}(w) = \|D[[w]] - Y\|_2^2,$$

where

$$[[w]] = (w_1^{(2)} w_1^{(1)\top}, w_2^{(2)} w_2^{(1)\top}, \ldots, w_n^{(2)} w_n^{(1)\top})^\top. \quad (8)$$

Therefore, we achieve a global minimum if $\hat{Y}$ is the projection of $Y$ onto the column space of $D$.

If $w_k^{(2)} \neq 0$ for all $k \in \{1, 2, \ldots, n\}$, the column space of $D'$ is equal to that of $D$. Hence, $\hat{Y}$ being the projection of $Y$ onto the column space of $D$ is achieved by our parameterization, which completes the proof of Theorem 2 (iv). Any critical point w.r.t. $(w^{(1)}, w^{(2)})$ needs to be a critical point w.r.t. $w^{(1)}$. Hence, every critical point is a global minimum if $w_k^{(2)} \neq 0$ for all $k \in \{1, 2, \ldots, n\}$. This completes the proof of Theorem 2 (iii). ■

**Proof of Theorem 2 (ii)** From Theorem 2 (iii), if $w_k^{(2)} \neq 0$ for all $k \in \{1, 2, \ldots, n\}$, every local minimum is a global minimum (because a set of local minima is included in a set of critical points).

Consider any point $\bar{w}$ where $\bar{w}_k^{(2)} = 0$ for some $k \in \{1, 2, \ldots, n\}$. Without loss of generality, with some integer $c$, let $\bar{w}_k^{(2)} \neq 0$ for all $k \in \{1, \ldots, c-1\}$ and $\bar{w}_k^{(2)} = 0$ for all $k \in \{c, \ldots, n\}$.

Then, at the point $\bar{w}$,

$$\mathcal{L}(\bar{w}) = \|D_{\cdot 1:c-1}[[\bar{w}]]_{1:c-1} - Y\|_2^2,$$

where $D_{\cdot 1:c-1}$ and $[[\bar{w}]]_{1:c-1}$ are the block elements of $D$ (equation 4) and $[[\bar{w}]]$ (equation 8) that correspond to the nonzero $\bar{w}_k$. Then, from the proof of Theorem 2 (iii), if $\bar{w}$ is a critical point, then $D_{\cdot 1:c-1}[[\bar{w}]]_{1:c-1}$ is the projection of $Y$ onto $D_{\cdot 1:c-1}$.

If $\bar{w}$ is a local minimum, for sufficiently small $\epsilon > 0$, for any $\bar{k} \in \{c, \ldots, n\}$, for any $(w_{\bar{k}}^{(1)}, w_{\bar{k}}^{(2)}) \in \mathbb{R}^d \times \mathbb{R}$ ,

$$\begin{aligned}&\|D_{\cdot 1:c-1}[[\bar{w}]]_{1:c-1} - Y\|_2^2 \\ &\leq \|D_{\cdot 1:c-1}[[\bar{w}]]_{1:c-1} + D_{\cdot \bar{k}}[[\bar{w}_\epsilon]]_{\bar{k}} - Y\|_2^2,\end{aligned}$$

which is simplified to

$$\begin{aligned}0 \leq& 2(D_{\cdot 1:c-1}[[\bar{w}]]_{1:c-1} - Y)^\top (D_{\cdot \bar{k}}[[\bar{w}_\epsilon]]_{\bar{k}}) \\ &+ \|D_{\cdot \bar{k}}[[\bar{w}_\epsilon]]_{\bar{k}}\|^2, \end{aligned} \tag{9}$$

where $[[\bar{w}_\epsilon]]_{\bar{k}}$ is a $\epsilon$-perturbation of $[[\bar{w}]]_{\bar{k}}$ as

$$[[\bar{w}_\epsilon]]_{\bar{k}} = \epsilon w_{\bar{k}}^{(2)} (\bar{w}_{\bar{k}}^{(1)} + \epsilon w_{\bar{k}}^{(1)}).$$

Here, where $D_{\cdot,\bar{k}}$ is the corresponding $\bar{k}$-th block of size $m \times d$. In equation (9), the first term contains $\epsilon$ and $\epsilon^2$ terms whereas the second term contains $\epsilon^2$, $\epsilon^3$ and $\epsilon^4$ terms. With $\epsilon$ sufficiently small, the $\epsilon$ term must be zero (as we can change its sign by the direction of perturbation). Then, with $\epsilon$ sufficiently small, the $\epsilon^2$ terms become dominant and must satisfy

$$\begin{aligned}0 \leq& 2\epsilon^2(D_{\cdot 1:c-1}[[\bar{w}]]_{1:c-1} - Y)^\top (D_{\cdot \bar{k}} w_{\bar{k}}^{(1)} w_{\bar{k}}^{(2)}) \\ &+ \epsilon^2 \|D_{\cdot \bar{k}} \bar{w}_{\bar{k}}^{(1)} w_{\bar{k}}^{(2)}\|^2,\end{aligned}$$

for any $(w_{\bar{k}}^{(1)}, w_{\bar{k}}^{(2)}) \in \mathbb{R}^d \times \mathbb{R}$, which implies

$$0 \leq 2\epsilon^2(D_{\cdot 1:c-1}[[\bar{w}]]_{1:c-1} - Y)^\top (D_{\cdot \bar{k}} w_{\bar{k}}^{(1)} w_{\bar{k}}^{(2)}),$$

for any $|w_{\bar{k}}^{(2)}| << \min_{j \in \{1,2,\ldots,d\}} |(w_{\bar{k}}^{(1)})_j|$ (see footnote[3]). It implies that for any $\bar{k} \in \{c, \ldots, n\}$,

$$0 = (D_{\cdot 1:c-1}[[\bar{w}]]_{1:c-1} - Y)^\top D_{\cdot \bar{k}}.$$

Since $D_{\cdot 1:c-1}[[\bar{w}]]_{1:c-1}$ is the projection of $Y$ onto $D_{\cdot 1:c-1}$ (as discussed above), this implies that $D^\top(D[[\bar{w}]] - Y) = (D_{\cdot 1:c-1}[[\bar{w}]]_{1:c-1} - Y)^\top D = 0$. Thus, $D[[\bar{w}]]$ is the projection of $Y$ onto $D$, which is a global minimum. ■

[3]For readers who have experienced deriving some conditions of a local minimum, it may sound too good to be true that we can ignore the $\epsilon^2$ term from $\|D_{\cdot \bar{k}}[[\bar{w}_\epsilon]]_{\bar{k}}\|^2$. However, this is true and not that good since we are considering a spacial case of $\bar{w}_{\bar{k}}^{(2)} = 0$. Indeed, if $\bar{w}_{\bar{k}}^{(2)} \neq 0$, $[[\bar{w}_\epsilon]]_{\bar{k}} = (\bar{w}_{\bar{k}}^{(2)} + \epsilon w_{\bar{k}}^{(2)})(\bar{w}_{\bar{k}}^{(1)} + \epsilon w_{\bar{k}}^{(1)})$ creates the $\epsilon^2$ term from $\|D_{\cdot \bar{k}}[[\bar{w}_\epsilon]]_{\bar{k}}\|^2$ that we cannot ignore.

**Proof of Theorem 2 (i)** It is sufficient to prove non-convexity w.r.t. the parameters $(w_k^{(1)}, w_k^{(2)})$ for each $k \in \{1, 2, \ldots, n\}$; that is, we prove that $\mathcal{L}'(w_1^{(k)}, w_k^{(2)}) = \mathcal{L}(w)|_{w_{i \neq k}=0} = \|D_{\cdot k} w_k^{(1)} w_k^{(2)} - Y\|_2^2$ is non-convex, where $D_{\cdot,k}$ is the corresponding $k$-th block of size $m \times d$. It is indeed easy to see that $\mathcal{L}'$ is non-convex. For example, at $(w_k^{(1)}, w_k^{(2)}) = 0$, its Hessian is

$$\begin{bmatrix} 0 & D_{\cdot k} \\ D_{\cdot k} & 0 \end{bmatrix} \not\succeq 0,$$

if $D_{\cdot k} \neq 0$. Thus, it is non-convex if $D \neq 0$. ■

## Proof of Theorem 3 (Generalization Bound for Shallow Model)

With a standard use of Rademacher complexity (e.g., see section 3 in Mohri, Rostamizadeh, and Talwalkar for a clear introduction of Rademacher complexity, and the proof of lemma 12 in Xie, Liang, and Song for its concrete use),

$$\begin{aligned}&\frac{1}{2}\mathbb{E}_x(f(x) - \hat{f}(x; w^*))^2 - \mathcal{L}(w) \\ &\leq (C_Y^2 + C_{\hat{Y}}^2)\sqrt{\frac{\log \frac{1}{\delta}}{2m}} + 2(C_Y + C_{\hat{Y}})\mathcal{R}_m(\mathcal{F}),\end{aligned} \tag{10}$$

with probability at least $1 - \delta$, where $(C_Y^2 + C_{\hat{Y}}^2)$ is an upper bound on the value of $\mathcal{L}(w)$, and $(C_Y + C_{\hat{Y}})$ comes from an upper bound on the Lipschitz constant of the squared loss function. Note that $f(x; w) = [\![\sigma, x]\!]^\top [\![w]\!] \leq C_W C_{\sigma x}$.

We first compute an upper bound on the empirical Rademacher complexity of our model class $\mathcal{F}$ as follows: with Rademacher variables $\xi_i$,

$$\begin{aligned} m\hat{\mathcal{R}}_m(\mathcal{F}) &= \mathbb{E}\left[\sup_{w \in S_w} \sum_{i=1}^m \xi_i [\![\sigma, x_i]\!]^\top [\![w]\!]\right] \\ &\leq \mathbb{E}\left[\sup_{w \in S_w} \left\|\sum_{i=1}^m \xi_i [\![\sigma, x_i]\!]\right\|_2 \|[\![w]\!]\|_2\right] \\ &= C_W \mathbb{E}\left[\left\|\sum_{i=1}^m \xi_i [\![\sigma, x_i]\!]\right\|_2\right] \end{aligned}$$

where the second line follows the Cauchy-Schwarz inequality. Furthermore,

$$\begin{aligned} \mathbb{E}\left[\left\|\sum_{i=1}^m \xi_i [\![\sigma, x_i]\!]\right\|_2\right] &\leq \sqrt{\mathbb{E}\left[\sum_{i=1}^m \sum_{j=1}^m \xi_i \xi_j [\![\sigma, x_i]\!]^\top [\![\sigma, x_j]\!]\right]} \\ &\leq \sqrt{\sum_{i=1}^m \|[\![\sigma, x_i]\!]\|_2^2} \\ &\leq C_{\sigma x}\sqrt{m}. \end{aligned}$$

where the first line uses Jensen's inequality for the concave function, and the second line follows that for any $z$, $\mathbb{E}\left[\sum_{j=1}^{m}\sum_{i=1}^{m}\xi_i\xi_j z_i z_j\right] \leq \sum_{i=1}^{m} z_i^2$, where we used the definition of Rademacher variables for $E[\xi_i\xi_j]$. Hence, $\hat{\mathcal{R}}_m(\mathcal{F}) \leq \frac{C_W C_{\sigma x}}{\sqrt{m}}$. Since the Rademacher complexity $\mathcal{R}_m(\mathcal{F})$ is the expectation of $\hat{\mathcal{R}}_m(\mathcal{F})$ over samples $(x, y)$, $R_m(\mathcal{F}) \leq \frac{C_W C_{\sigma x}}{\sqrt{m}}$.

■

### Tensorial Structure

Since the output of our network is the sum of the outputs of all the paths in the network, we observe the following interesting structure:

$$\hat{f}^s_{n_1,\ldots,n_H}(x) = \sum_{k_0=0}^{d}\sum_{k_1=1}^{n_1}\cdots\sum_{k_H=1}^{n_H} [\![\sigma]\!]_{k_1,\ldots,k_H}(x)\mathbf{x}_{k_0}[\![w]\!]_{k_0,k_1,\ldots,k_H},$$

where

$$[\![\sigma]\!]_{k_1,\ldots,k_H}(x) = \sigma_s(\mathbf{x}^\top\mathbf{r}^{(1)}_{k_1})\prod_{l=2}^{H}\sigma_s(h_r^{(l-1)}(x)\, r^{(l)}_{k_l}),$$

and $[\![w]\!]_{k_0,k_1,\ldots,k_H} = \left(\prod_{l=1}^{H} w^{(l)}_{k_{l-1}k_l}\right) w^{(H+1)}_{k_H}$. This means that the function is the weighted combination of $(d+1) \times n_1 \times n_2 \ldots n_H$ nonlinear basis functions, which is exponential in the number of layers $H$. Alternatively, the function can also be viewed as the inner product between two tensors $[\![\sigma]\!]_{k_1,\ldots,k_H}(x)\mathbf{x}_{k_0}$ and $[\![w]\!]_{k_0,k_1,\ldots,k_H}$, which is nonlinear in input $x$, but *linear* in the parameter tensor $[\![w]\!]_{k_0,k_1,\ldots,k_H}$. We note that the parameter tensor is not arbitrary but highly structured: it is composed using a collection of matrices with $d_w = (d+1)n_1 + n_H + \sum_{l=2}^{H} n_{l-1}n_l$ number of adjustable parameters. Such special structure allows the function to generate an exponential number of basis functions yet keep the parameterization compact.

## Proofs for Multilayer Model

In this section, we provide the proofs of the theorems and corollary presented in Section "Multilayer Model".

### Proof of Corollary 4 (Universal Approximation with Deep Model)

From the definition of $\mathbf{r}_1^{(1)}, r_1^{(2)}, r_1^{(3)}, \ldots, r_1^{(H)}$, the first unit in each hidden layer of the main semi-random net is not affected by the random net; the output of the first unit of semi-random net is multiplied by one. By setting most of the weights $w$ to zeros, we can use only the first unit at each layer from the second hidden layer, creating a single path that is not turned-off by the random activation net from the second hidden layer. Then, by adjusting weights $w$ in the path, we can create an identity map from the second hidden layer unit, from which the path starts. In theorem 1, we have already shown that the output of any second hidden layer unit has universal approximation ability. This completes the proof.

### Proof of Theorem 5 (Lower Bound on Universal Approximation Power)

Since the output of our network is the sum of the outputs of all the paths in the network, we observed in Section "Tensorial Structure" that

$$\hat{f}_{n_1,\ldots,n_H}(x) = \sum_{k_0=0}^{d}\sum_{k_1=1}^{n_1}\cdots\sum_{k_H=1}^{n_H} [[\sigma]]_{k_1,\ldots,k_H}(x)\mathbf{x}_{k_0}[[w]]_{k_0,k_1,\ldots,k_H},$$

where

$$[[\sigma]]_{k_1,\ldots,k_H}(x) = \sigma_s(\mathbf{x}^\top\mathbf{r}^{(1)}_{k_1})\prod_{l=2}^{H}\sigma_s(h_r^{(l-1)}(x) r^{(l)}_{k_l}),$$

and

$$[[w]]_{k_0,k_1,\ldots,k_H} = \left(\prod_{l=1}^{H} w^{(l)}_{k_{l-1}k_l}\right) w^{(H+1)}_{k_H}.$$

By re-writing $g_{k0,k1,\ldots,k_H}(x) = [[\sigma]]_{k_1,\ldots,k_H}(x)\mathbf{x}_{k_0}$, this means that

$$\mathcal{F}^s_{n_1,\ldots,n_H} \subseteq \mathrm{span}(\{g_{k0,k1,\ldots,k_H} : \forall k_0, \forall k_1, \ldots, \forall k_H\}).$$

Then, Theorem 5 is a direct application of the following result:

**Lemma 12** (Barron 1993, Theorem 6) *For any fixed set of $M$ functions $g_1, g_2, \ldots, g_M$,*

$$\sup_{f\in\Gamma_C}\ \inf_{\hat{f}\in\mathrm{span}(g_1,g_2,\ldots,d_M)} \|f-\hat{f}\|_{L^2(\Omega)} \geq \kappa\frac{C}{d}M^{-1/d}.$$

We apply Lemma 12 with our $g_{k0,k1,\ldots,k_H}$ to obtain the statement of Theorem 5.

■

### Proof of Corollary 6 (No Bad Local Minima and Few Bad Critical Points w.r.t. Two Last Layers)

It directly follows the proof of Theorem 2, because any critical point (or any local minimum) is a critical point (or respectively, a local minimum) with respect to $(W^{(H)}, W^{(H+1)})$. ■

### Proof of Corollary 7 (Generalization Bound for Deep Model)

By noticing that $f_{n_1,\ldots,n_H}(x) = vec([\![w]\!])^\top vec([\![\sigma, x]\!])$ (i.e., the tensorial structure as discussed in the proof of Theorem 5), it directly follows the proof of Theorem 3 (Generalization Bound – shallow); in the proof of Theorem 3, replace $[\![\sigma, x_i]\!]$ and $[\![w]\!]$ with the corresponding definitions for multilayer models, which are $vec([\![\sigma, x_i]\!])^\top$ and $vec([\![w]\!])$ of multilayer models.

■

## Discussion on an Upper Bound on Approximation Error for Multilayer Model

Here, we continue the discussion in Section "Benefit of Depth". An upper bound on approximation error can be decomposed into three illustrative terms as

$$\begin{aligned} &\|f - \hat{f}\|^2_{L^2(\Omega)} \\ &= \sum_{i \in S} \|f - \hat{f}\|^2_{L^2(\Omega_i)} \qquad (11) \\ &\leq \sum_{i \in S} \|f - f_i\|^2_{L^2(\Omega_i)} + \|\hat{f} - \hat{f}_i\|_{L^2(\Omega_i)} + \|f_i - \hat{f}_i\|^2_{L^2(\Omega_i)}, \end{aligned}$$

where $\{\Omega_i\}_{i \in S}$ is an arbitrary partition of $\Omega$, and $f_i = f(x_i)$ is a constant function with some $x_i \in \Omega_i$. Then, the first term and the second term represent how much each of $f$ and $\hat{f}$ varies in $\Omega_i$. In other words, we can bound these two terms by some "smoothness" of $f$ and $\hat{f}$ (some discontinuity at a set of measure zero poses no problem as we are taking the norm of $L^2(\Omega_i)$ space). The last term in equation (11) represents the expressive power of $\hat{f}$ on the finite points $\{x_i\}_{i \in S}$ (assuming that $S$ is finite). Essentially, to obtain a good upper bound, we want to have a $\hat{f}$ that is "smooth" and yet expressive on some finite points.

For the first term in equation (11), we can obtain a bound via a simple calculation for any $f \in \Gamma_C$ as follows: by the mean value theorem, there exists $z_x$ for each $x$ such that

$$\begin{aligned} \|f - f_i\|^2_{L^2(\Omega_i)} &= \int_{x \in \Omega_i} (f(x) - f(x_i))^2 \\ &= \int_{x \in \Omega_i} (\nabla f(z_x)^\top (x - x_i))^2 dx \\ &\leq \|\nabla f(z_x)\|^2_{L^2(\Omega_i)} \|x - x_i\|^2_{L^2(\Omega_i)} \\ &\leq \|\nabla f(x)\|^2_{L^2(R^d)} \|x - x_i\|^2_{L^2(\Omega_i)} \\ &\leq C^2 \|x - x_i\|^2_{L^2(\Omega_i)}. \end{aligned}$$

where $\|x - x_i\|^2_{L^2(\Omega_i)} = \int_{x \in \Omega_i} \|x - x_i\|^2_2$. Hence, the bound goes to zero as the size of $\Omega_i$ decreases.

We can use the same reasoning to bound the second term in equation (11) with some "smoothness" of $\hat{f}$. A possible concern is that $\hat{f}$ becomes less "smooth" as the width $n$ and depth $H$ increase. However, from Bessel's inequality with Gram-Schmidt process, it is clear that such an effect is bounded for a best $\hat{f}$ with a finite $s \geq 1$ (notice the difference from statistical learning theory, where we cannot focus on the best $\hat{f}$). The last term in equation (11) represents the error at points $x_i \in \Omega_i$, which would be bounded via optimization theory, as the expressive power of $\hat{f}$ increases as depth $H$ and width increase.

While this reasoning illustrates what factors may matter, a formal proof is left to future work.

## Additional Experimental Details

In this section, we provide additional Experimental details.

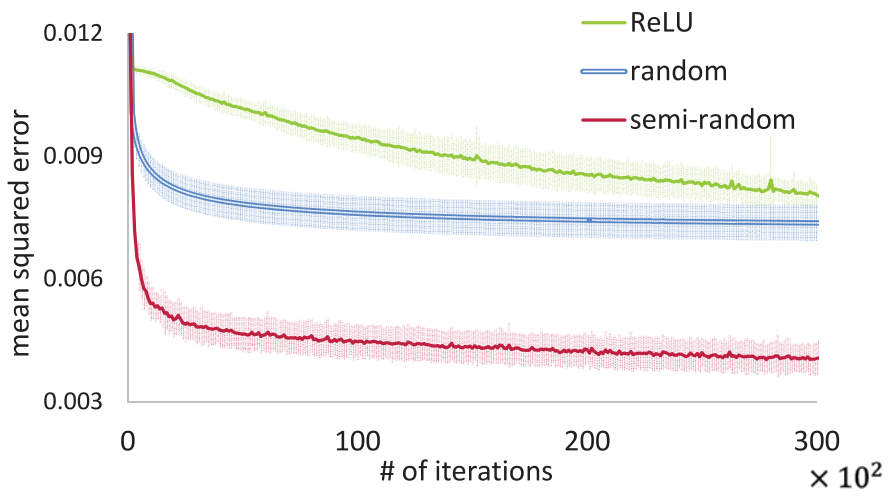

Figure 5: Training error for a simple test function.

### A simple test function

Figure 5 shows the training errors for a sine function experiment discussed in Section "Simple Test Function". It shows roughly the same patterns as in their test errors, indicating that there is no large degree of overfitting.

Figures 6, 7, and 8 visualize the function learned at each iteration for each method. We can see that semi-random features (LSR) learns the function quickly.

At each trial, $5000$ points of the inputs $x$ were sampled uniformly from $[-12\pi, 12\pi]$ for each of training dataset and test dataset. We repeated this trial $20$ times to obtain standard deviations. All hyperparameters were fixed to the same values for all methods, with the network architecture being fixed to 1-50-50-1. For all the methods, we used the learning rate $5 \times 10^{-4}$, momentum parameter $0.9$, and mini-batch size of $500$. For all methods, the weights are initialized as $0.1$ times random samples drawn from the standard normal distribution.

### UCI datasets

For all methods on all datasets, we use SGD with 0.9 momentum, a batch size of 128 to run 100 epochs (passes over the whole dataset). The initial learning rate is set to 0.1 (for some combinations, this leads to NaN and we use 0.02 as initial learning rate). We also use an exponential staircase decaying schedule for the learning rate where after one epoch we decrease it to 0.95 of the original rate. The parameters are initialized as normal distributions times $1/\sqrt{d}$ where $d$ is the input dimension for the layer.

For ReLU and semi-random features (LSR and SSR), their behaviors change similarly on most datasets: more layers and more units lead to better performance. Also, on most datasets, adding more layers leads to better performance than adding more units per layer. The dataset adult is an exception where all methods have similar performance and more parameters actually lead to worse performance. This is likely due to that adult is quite noisy and using more parameters leads to overfitting. Overfitting is also observed on senseit dataset for LSR where the training errors keep become smaller but test errors increase.

### Image datasets

We train these convolution neural networks using SGD with 0.9 momentum with a batch size of 64 for MNIST and 128

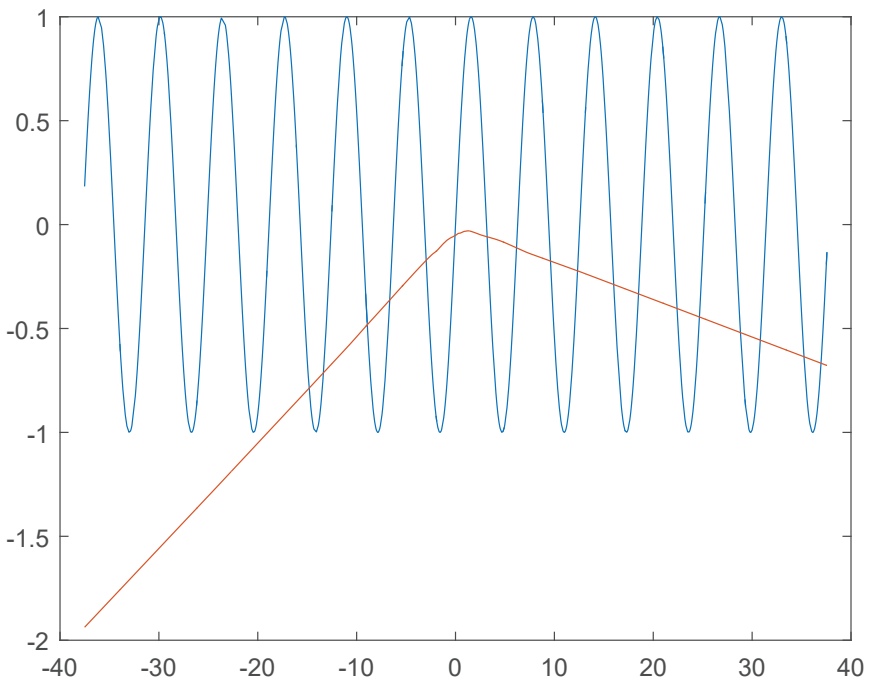

(a) iteration 0

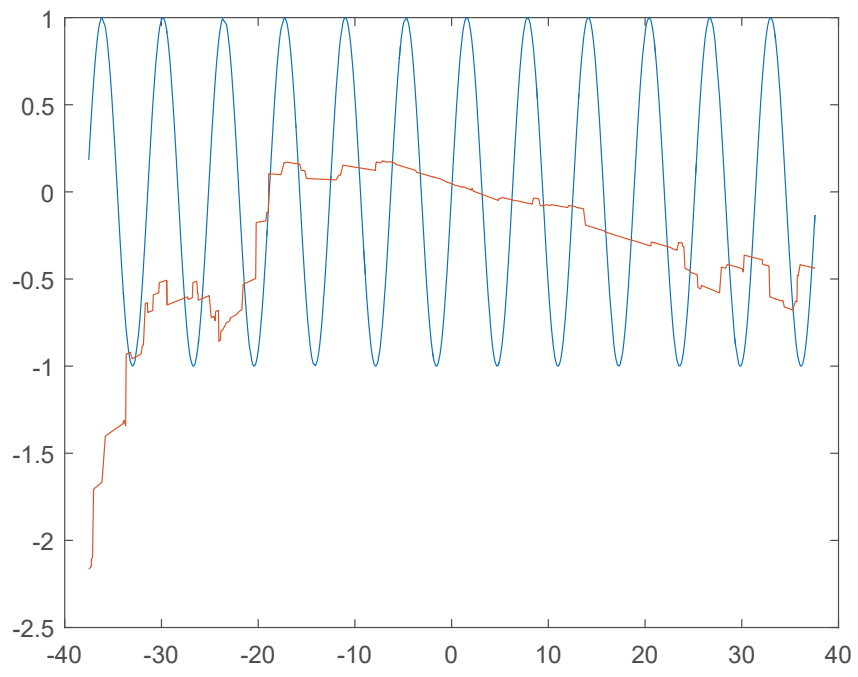

(b) iteration $10^2$

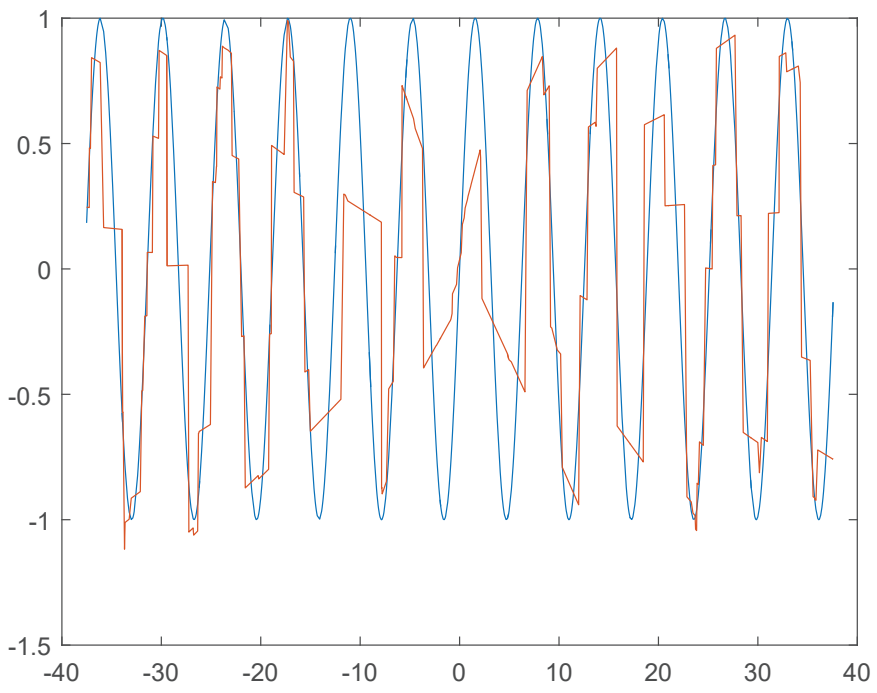

(c) iteration $10^3$

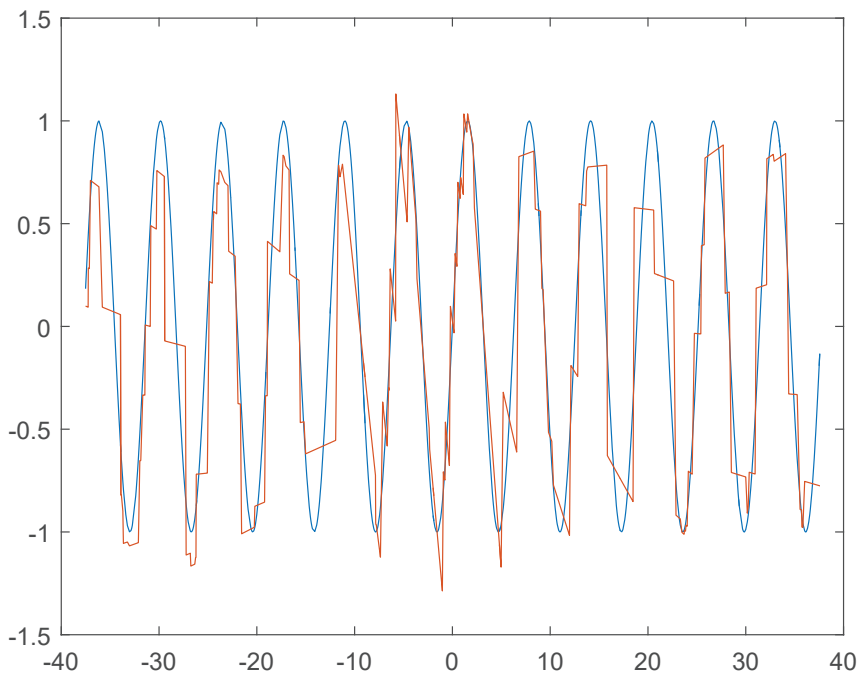

(d) iteration $10^4$

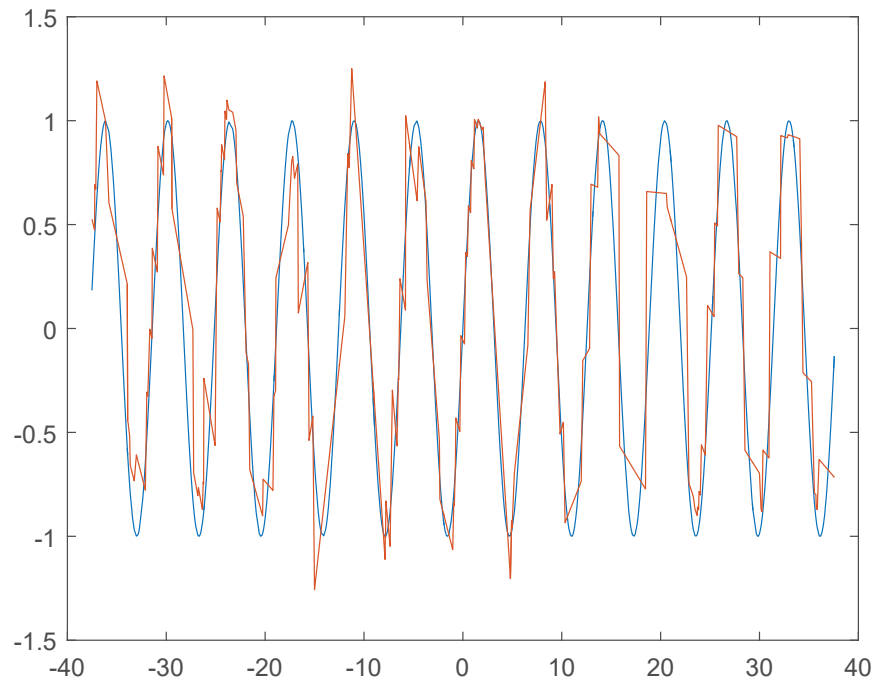

(e) iteration $5 \times 10^4$

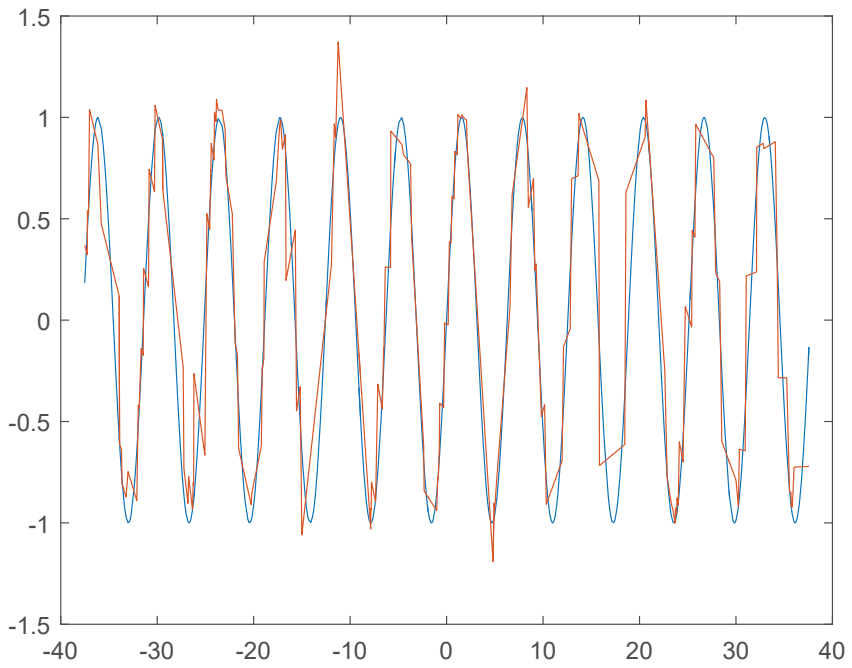

(f) iteration $10^5$

Figure 6: Function learned at each iteration (Semirandom - LSR).

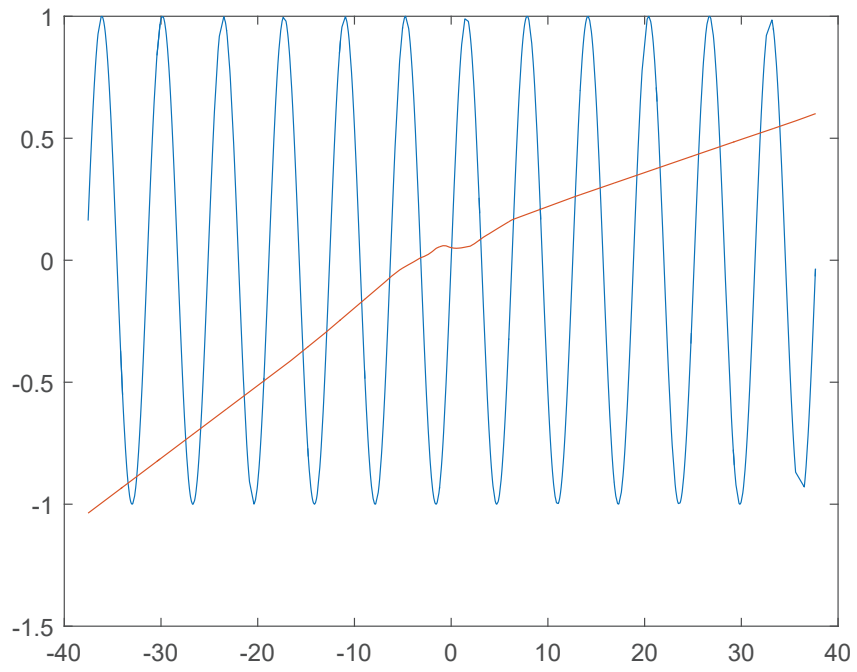

(a) iteration 0

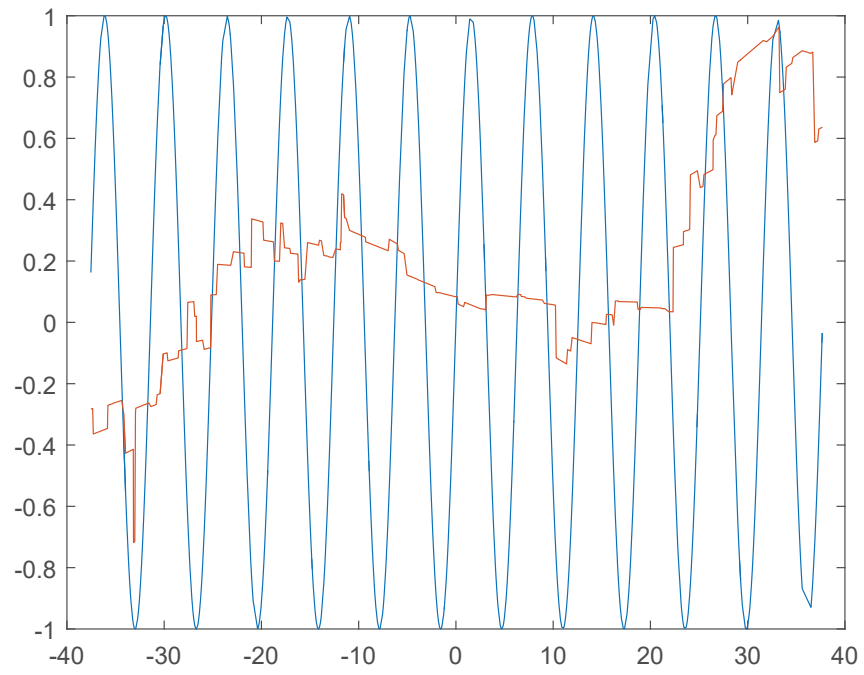

(b) iteration $10^2$

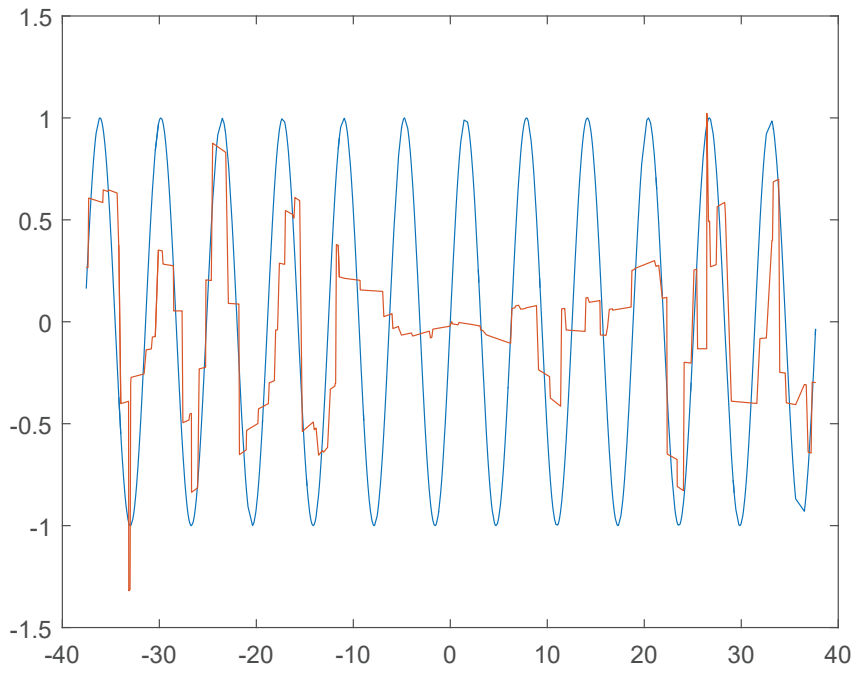

(c) iteration $10^3$

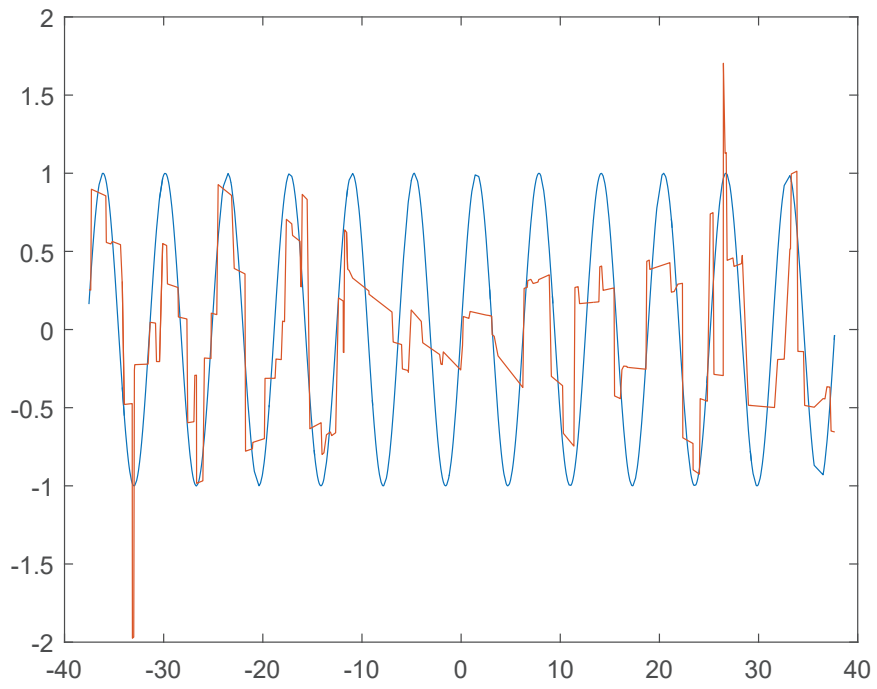

(d) iteration $10^4$

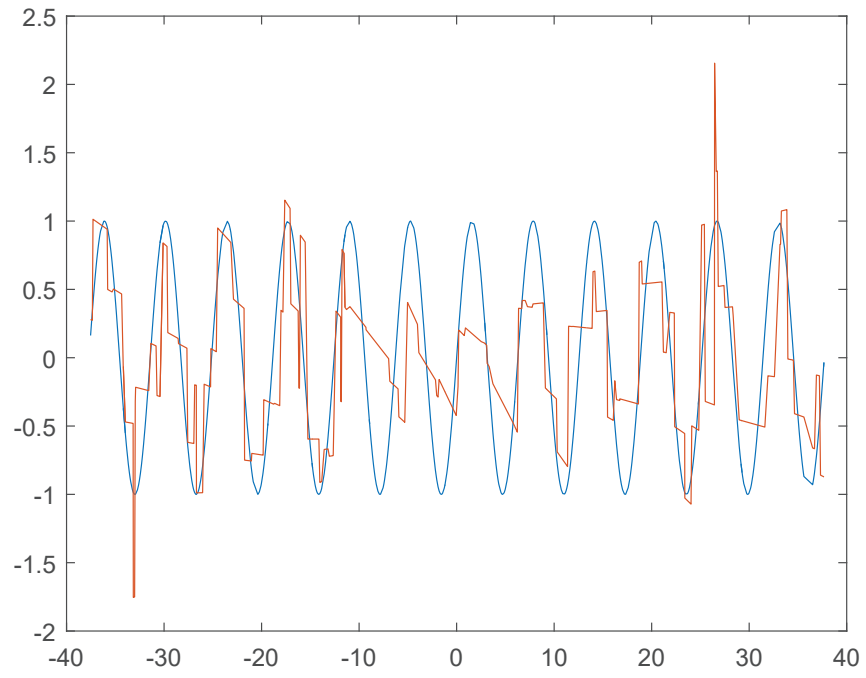

(e) iteration $5 \times 10^4$

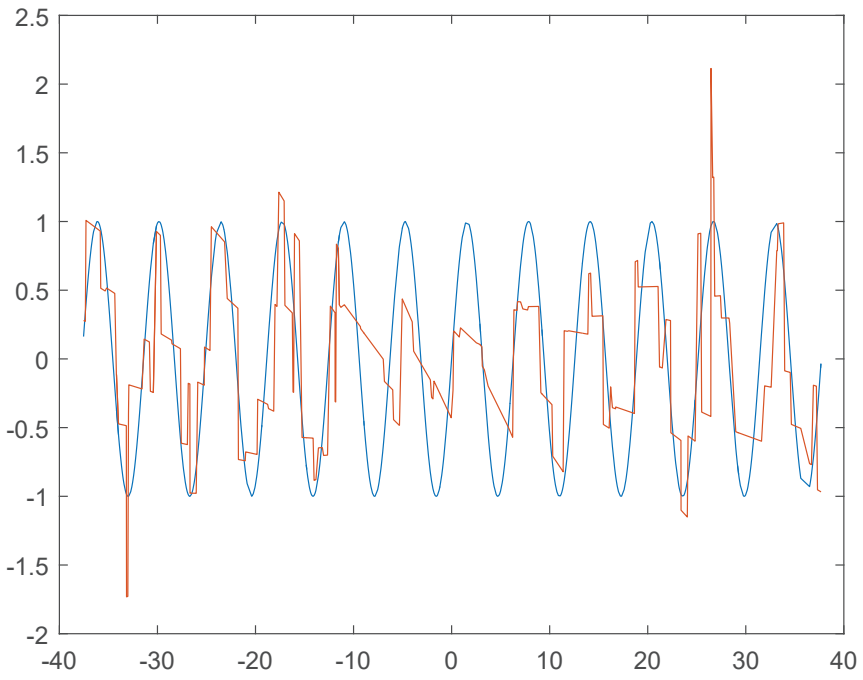

(f) iteration $10^5$

Figure 7: Function learned at each iteration (fully random).

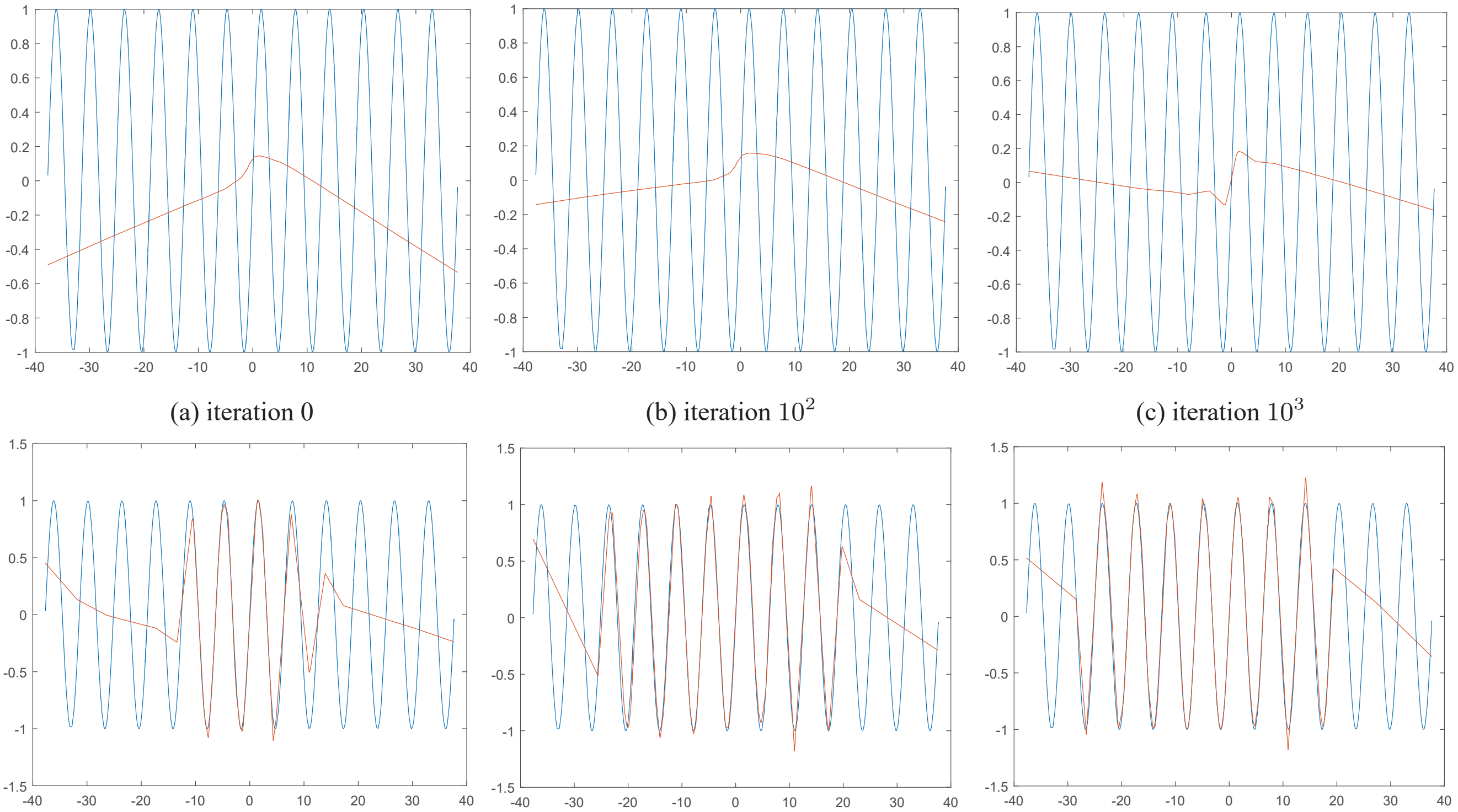

(a) iteration 0

(b) iteration $10^2$

(c) iteration $10^3$

(d) iteration $10^4$

(e) iteration $5 \times 10^4$

(f) iteration $10^5$

Figure 8: Function learned at each iteration (ReLU).

for CIFAR10 and SVHN. On each dataset, we have tried initial learning rates of 0.1, 0.01, and 0.001 and used an exponential decaying schedule. For MNIST, the decay schedule is decreasing to 0.95 of the previous rate after each epoch. For CIFAR10 and SVHN, the schedule is decreasing to 0.1 of the previous rate after 120 epochs. The best performing results on a validation set are then picked as the final solution, and usually the learning rate of 0.1 worked best. The parameters are also randomly initialized from normal distributions times $1/\sqrt{d}$ where $d$ is the input dimension for the layer.

## Better than Random Feature?

The experimental results verified our intuition that semi-random feature can outperform random feature with fewer number of unites due to its learnable weights. We can also strengthen this intuition via the following theoretical insights. Let $f \in \mathcal{F}_{\text{random},n_1 \cdots n_H}$ be a function that is a composition of any fully-random features with depth $H$ where the adjustable weights are only in the last layer. The following corollary states that a model class of any fully-random features has a approximation power exponentially bad in the dimensionality of $x$ in the worst case.

**Corollary 13** (Lower bound on approximation power for fully-random feature) *Let $\Omega = [0,1]^d$. For any depth $H \geq 0$, and for any set of nonzero widths $\{n_1, n_2, \ldots, n_H\}$,*

$$\sup_{f \in \Gamma_C} \inf_{\hat{f} \in \mathcal{F}_{\text{random},n_1 \cdots n_H}} \|f - \hat{f}\|_{L^2(\Omega)} \geq \frac{\kappa C}{d^2} (n_H)^{-1/d},$$

*where $\kappa \geq (8\pi e^{(\pi-1)})^{-1}$ is a constant.*

**Proof** The model class of any random features is the span of the features represented at the last hidden layer. Hence, the statement of the corollary directly follows from the proof of Theorem 5. ■

Corollary 13 (lower bound for fully-random feature) together with Theorem 5 (lower bound for semi-random feature) reflects our intuition that semi-random feature model can potentially get exponential advantage over random feature by learning hidden layer's weights. Again, because the lower bound may not be tight, this is intended only to aid our intuition.

We can also compare upper bounds on their approximation errors with an additional assumption. Assume that we can represent a target function $f$ using some basis as

$$f(x) = \int_{r \in \mathbb{S}^{d-1}, \|w\| \leq C_W} \sigma(r^\top x)(w^\top x)\, p(r, w).$$

Then, we can obtain the following results.

- If we have access to the true distribution $p(r, w)$, $f(x)$ can be approximated as a finite sample average, obtaining approximation error of $O(\frac{1}{\sqrt{n}})$.

- Without knowing the true distribution $p(r, w)$, a purely random feature approximation of $f(x)$ incurs a large approximation error of $O(\frac{c}{q_0 \cdot q_1 \cdot \sqrt{n}})$. Here, $q_0$ is inverse of the hyper-surface area of a unit hypersphere and $q_1$ is the inverse of the volume of a ball of radius $C_W$.
- Without knowing the true distribution $p(r, w)$, semi-random feature approach with one hidden layer model can obtain a smaller approximation error of $O(\frac{35c}{\sqrt{n}})$.

A detailed derivation of this result is presented in Appendix.

## Derivation of Upper Bounds on Approximation Errors

With an additional assumption, we compare upper bounds on approximation errors for random features and semi-random features. Recall the additional assumption that we can represent a target function $f$ using some basis as

$$f(x) = \int_{r \in \mathbb{S}^{d-1}, \|w\| \le C_W} \sigma(r^\top x)(w^\top x)\, p(r, w),$$

where we can write $p(r, w) = p(r)\, p(w|r)$. If we have access to the true distribution $p(r, w)$, then $f(x)$ can be approximated as a finite sample average

$$\widehat{f}_1(x) = \frac{1}{n} \sum_{i=1}^{n} \sigma(r_i^\top x)(w_i^\top x)$$

$$(r_i, w_i) \overset{i.i.d.}{\sim} p(r, w).$$

Such an approximation will incur an error of $O(\frac{1}{\sqrt{n}})$.

On the other hand, without knowing the true distribution $p(r, w)$, a purely random feature approximation of $f(x)$ will be sampling both $r$ and $w$ from uniform distribution: with $r_i \overset{i.i.d.}{\sim} q(r) :=$ Uniform $(\mathbb{S}^{d-1})$ and $w_i \overset{i.i.d.}{\sim} q(w) :=$ Uniform $(\|w_i\| \le C_W)$,

$$\begin{aligned}\widehat{f}_2(x) &= \frac{1}{n} \sum_{i=1}^{n} \frac{p(r_i, w_i)}{q(r_i) q(w_i)} \sigma(r_i^\top x)(w_i^\top x) \\ &= \frac{1}{n} \sum_{i=1}^{n} \alpha_i\, \sigma(r_i^\top x)(w_i^\top x)\end{aligned}$$

where $q(r) = q_0 := \Gamma(\frac{n}{2})/(2\pi^{\frac{d}{2}})$ (inverse of the hyper-surface area of a unit hypersphere) and $q(w) = q_1 := \Gamma(\frac{d}{2} + 1)/(\pi^{\frac{d}{2}} C_W^d)$ (inverse of the volume of a ball of radius $C_W$). Interestingly enough, for the unit hypersphere, the hyper-surface area reaches a maximum and then decreases towards 0 as $d$ increases. One can show that the seven-dimensional unit hypersphere has a maximum hyper-surface area that is less than 35. But the volume of a ball of radius $C_W$ depends exponentially on the radius $C_W$. If $\|p(r, w)\|_\infty = c$, then the importance weight $\alpha_i$ can be as large as $\frac{c}{q_0 \cdot q_1}$. Due to the fact that $q_1$ can be very small which make the bound very large, this incurs a big approximation error of $O(\frac{c}{q_0 \cdot q_1 \cdot \sqrt{n}})$.

In contrast, if we sample $r$ from unit sphere and then optimize over $w$ (i.e., semi-random approach with one hidden layer model), we obtain the following approximation: with $r_i \overset{i.i.d.}{\sim} q(r) :=$ Uniform $(\mathbb{S}^{d-1})$,

$$\begin{aligned}\widehat{f}_3(x) &= \frac{1}{n} \sum_{i=1}^{n} \frac{p(r_i)}{q(r_i)} \sigma(r_i^\top x)(w_i^\top x) \\ &= \frac{1}{n} \sum_{i=1}^{n} \beta_i\, \sigma(r_i^\top x)(w_i^\top x).\end{aligned}$$

By finding the best $w$, we can get at least as good as the case of $w_i \overset{i.i.d.}{\sim} p(w|r)$ where $p(w|r) = p(w, r)/p(r)$ is the true conditional distribution given $r$. If $\|p(r)\|_\infty = c$, then the important weight $\beta_i$ can be much smaller, with a bound of $\frac{c}{q_0} \le 35c$ independent of the dimension. Accordingly, this approximation incurs an error of $O(\frac{35c}{\sqrt{n}})$.

## Additional Future Work and Open Problems

We have discussed several open questions in the relevant sections of the main text. Here, we discusses additional future work and open problems.

We emphasize that our theoretical understanding of multilayer model is of preliminary nature and it is not intended to be complete. Future work may adopt an oscillation argument in the previous work by (Telgarsky 2016), in order to directly compare deep models and shallow models. To obtain a tight upper bound, future work may start with a stronger assumption such as one used in the previous work by (Rahimi and Recht 2009); an assumption on a relationship of random sampling and the true distribution. Indeed, when compared with neural network, semi-random feature would be advantageous when the target function is known to be decomposed into the unknown part $w$ and the random part **r** *with known distribution*. However, we did not assume such a knowledge in this paper, leaving the study of such a case to feature work.